%% file: main-CameraReady.tex
\documentclass[journal]{IEEEtai}

\usepackage{bm}
\usepackage{color,array}
\usepackage{graphicx}
\usepackage{textcomp}
\usepackage{stfloats}
\usepackage{url}
\usepackage{verbatim}
\usepackage{graphicx}
\usepackage{multirow}
\usepackage{xpatch}
\usepackage{tikz}
\usepackage{balance}
\usetikzlibrary{fit}
\usetikzlibrary{arrows}
\usetikzlibrary{shapes}
\usepackage[edges]{forest}
\usetikzlibrary{shadows.blur}
\usepackage{booktabs}
\usepackage{adjustbox}
\usepackage{amsmath,amssymb}
\usepackage{makecell}
\usepackage{bbm}
\usepackage{algorithm}  
\usepackage{algcompatible}
\usepackage{algorithmicx,algpseudocode}\usepackage{graphicx}
\usepackage{tikz,balance}

\usepackage{graphicx,tabularx, amsmath,algorithm,algpseudocode,algorithmicx}
\usepackage{amsfonts,amssymb, multirow}
\usepackage{url,makecell,balance,enumerate,tikz}
\usepackage{xcolor,booktabs,colortbl}
\usepackage{threeparttable}
\usepackage{cite}
\ifCLASSOPTIONcompsoc
  \usepackage[caption=false,font=normalsize,labelfont=sf,textfont=sf]{subfig}
\else
  \usepackage[caption=false,font=footnotesize]{subfig}
\fi

\def\NoNumber#1{{\def\alglinenumber##1{}\State #1}\addtocounter{ALG@line}{-1}}

\newcommand{\expect}{\mathbb{E}}

\def\NoNumber#1{{\def\alglinenumber##1{}\State #1}\addtocounter{ALG@line}{-1}}

\newcommand\blfootnote[1]{
    \begingroup
    \renewcommand\thefootnote{}\footnote{#1}
    \addtocounter{footnote}{-1}
    \endgroup
}

\makeatletter
\ExplSyntaxOn
\cs_new:Npn \bibColoredItems #1#2
  {
    \clist_map_inline:nn {#2} { \cs_new:cpn {bib@colored@##1} {#1} } 
  }
\ExplSyntaxOff

\newcommand\bib@setcolor[1]{%
  \ifcsname bib@colored@#1\endcsname
    \expandafter\color\expandafter{\csname bib@colored@#1\endcsname}
  \else
    \normalcolor
  \fi
}

\xpatchcmd\@bibitem
  {\item}
  {\bib@setcolor{#1}\item}
  {}{\PatchFailed}

\xpatchcmd\@lbibitem
  {\item}
  {\bib@setcolor{#2}\item}
  {}{\PatchFailed}
\makeatother

\begin{document}

\title{Expanding Horizons of Level Diversity via Multi-objective Evolutionary Learning}

\author{
Qingquan Zhang,~\IEEEmembership{Member,~IEEE},
Ziqi Wang~\IEEEmembership{Member,~IEEE},
Yuchen Li,~\IEEEmembership{Student Member,~IEEE},\\
Keyuan Zhang,~\IEEEmembership{Student Member,~IEEE},
Bo Yuan,~\IEEEmembership{Member,~IEEE},
Jialin Liu,~\IEEEmembership{Senior Member,~IEEE}
\thanks{ 
Q. Zhang, Z. Wang and B. Yuan are with the Guangdong Provincial Key Laboratory of Brain-inspired Intelligent Computation, Department of Computer Science and Engineering, Southern University of Science and Technology (SUSTech), Shenzhen, China.
}
\thanks{Y. Li is with the Tandon School of Engineering, New York University, USA. She contributed to this work when she was with the SUSTech.}
\thanks{K. Zhang is with the Virginia Polytechnic Institute and State University, USA. He contributed to this work when he was with the SUSTech.}
\thanks{J. Liu is with the School of Data Science, Lingnan University, Hong Kong SAR, China.}
\thanks{Corresponding author: Jialin Liu (jialin.liu@ln.edu.hk).}
}

\markboth{Journal of IEEE Transactions on Artificial Intelligence, Vol. 00, No. 0, Month 2020}
{First A. Author \MakeLowercase{\textit{et al.}}: Bare Demo of IEEEtai.cls for IEEE Journals of IEEE Transactions on Artificial Intelligence}

\maketitle

\begin{abstract}
In recent years, the generation of diverse game levels has gained increasing interest, contributing to a richer and more engaging gaming experience. A number of level diversity metrics have been proposed in literature, which are naturally multi-dimensional, leading to conflicted, complementary, or both relationships among these dimensions. However, existing level generation approaches often fail to comprehensively assess diversity across those dimensions. This paper aims to expand horizons of level diversity by considering multi-dimensional diversity when training generative models. We formulate the model training as a multi-objective learning problem, where each diversity metric is treated as a distinct objective. Furthermore, a multi-objective evolutionary learning framework that optimises multiple diversity metrics simultaneously throughout the model training process is proposed. Our case study on the commonly used benchmark \emph{Super Mario Bros.} demonstrates that our proposed framework can enhance multi-dimensional diversity and identify a Pareto front of generative models, which provides a range of tradeoffs among playability and two representative diversity metrics, including a content-based one and a player-centered one. Such capability enables decision-makers to make informed choices when selecting generators accommodating a variety of scenarios and the diverse needs of players and designers.~\blfootnote{The article has been accepted by IEEE Transactions on Artificial Intelligence, with the DOI: {10.1109//TAI.2024.3489534}.} ~\blfootnote{Code and results are available at \url{https://github.com/SUSTechGameAI/MultiDiversity_MOEL}.}

\end{abstract}

\begin{IEEEImpStatement}
Artificial intelligence-generated content (AIGC) techniques offer a new paradigm of content creation and have numerous applications in several industry sectors, including digital games. Evaluating game levels is crucial and should consider different aspects, with diversity being one of the most important. Multiple content-based and player-centered metrics have been proposed for measuring level diversity. However, their relationship is complex, and most existing works only consider one of them. Our proposed framework addresses this by considering multi-dimensional diversity and quality while training level generators. By providing game designers and developers with the ability to navigate various tradeoffs among multiple dimensions of diversity, our framework empowers them to tailor content to suit diverse player preferences, thereby enhancing player engagement and satisfaction. Additionally, our approach is not limited to multi-dimensional diversity and can accommodate other objectives, further enriching AIGC’s applicability. For instance, by incorporating additional objectives such as the degree of thrill or novelty, our framework can cater to a wider range of application scenarios, thus advancing the field of AIGC.

\end{IEEEImpStatement}

\begin{IEEEkeywords}
Artificial intelligence generated content, procedural content generation, content diversity, multi-objective evolutionary learning, games.
\end{IEEEkeywords}

\section{Introduction}

\IEEEPARstart{p}{rocedural} content generation (PCG) has been revolutionalised by the advances of machine learning over the past few years~\cite{shaker2016procedural,liu2021deep,guzdial2022procedural}. It can significantly contribute to reducing the burden on human designers and enhancing creativity in game design. Learning-based generators, such as generative adversarial nets (GANs)~\cite{10.1145/3422622} and variational autoencoders (VAEs)~\cite{DBLP:journals/corr/KingmaW13}, have demonstrated remarkable ability in generating game levels~\cite{volz2018evolving,snodgrass2020multi}, further amplifying the impact of PCG on game design.

In the realm of game level generation, the interplay between playability and diversity is significant in shaping the overall player experience~\cite{ yannakakis2018artificial}. Diversity focuses on fostering creativity and innovation in game content creation. Lack of diversity can lead to boredom and diminish player engagement~\cite{ yannakakis2018artificial, Liapis2019Orchestrating, gravina2019procedural}. Although there is no universally accepted metric for quantifying diversity~\cite{Jordanous2011Evaluating,liy2024measuring}, it can be conceptualised as multi-dimensional, considering both content-based and player-centered metrics~\cite{liy2024measuring}. These diversity metrics may conflict with, complement to each other, or both, offering a comprehensive framework for evaluating game level diversity. 

Playability and diversity metrics play crucial roles in guiding the generation of high-quality game levels as well. While the playability of generated content is often assured~\cite{volz2018evolving,shu2021experience,wang2022fun,Liapis2015Constrained,fontaine2021illuminating}, there has been growing research interest in generating diverse levels~\cite{ bontrager2021learning,preuss2014searching,gravina2019procedural,nam2019generation,beukman2022procedural,fontaine2021illuminating, lehman2011abandoning, sudhakaran2024mariogpt, Sarkar2024Procedural, AwiSch2023worldgan, torrado2020bootstrapping,wang2022fun}. To enhance the diversity of game levels, the playability and diversity are often converted into a single objective using the weighted sum method~\cite{preuss2014searching, fontaine2021illuminating,bontrager2021learning}. Studies like~\cite{preuss2014searching} and~\cite{fontaine2021illuminating} maintain a level archive, and continuously add new levels to increase diversity. Additionally, some works also focus on constructing generators to model a diverse level space. For example, Bontrager \textit{et al.}~\cite{bontrager2021learning} design a diversity loss, where diversity and playability terms are summarised into a unified loss using the weighted sum method to update the parameters of generators.

However, current studies, such as~\cite{preuss2014searching,bontrager2021learning,fontaine2021illuminating}, on improving generator diversity\footnote{In this paper, we refer to the generator's ability to generate diverse game levels as ``generator diversity''.} face the following three main challenges. 
\begin{itemize}
    \item 
    Challenge 1: Existing works often consider one single diversity metric, which potentially results in worse overall multi-dimensional diversity and negatively impacts user experience.

    \item Challenge 2: A few works consider diversity and playability metrics by taking a weighted sum, however, the process of determining appropriate weight values, presents a considerable challenge. These weight values are crucial as they emphasise the importance of various metrics when converting multiple objectives into a single one. The process of identifying a suitable set of weight values to align the trained model with decision-makers' specific preferences is often time-consuming.

    \item Challenge 3: Each run of the training process typically results in only one generator with consideration of one tradeoff predefined by weights. Due to conflicts among playability and multiple diversity metrics, it is impossible for a single tradeoff to simultaneously reach optimal values of playability and multiple diversity metrics. Changes in tradeoff requirements acquire additional costs to identify new suitable weights for training a new model.
\end{itemize}
Consequently, there is a need for a set of different tradeoffs among these metrics. The diverse tradeoffs empower game designers or players to tailor levels to suit diverse player preferences and scenarios.

We recognise the necessity of designing a model training framework that can simultaneously consider multi-dimensional diversity metrics without relying on the determination of weights to overcome the aforementioned challenges. Such a framework should also provide a diverse set of generators with different tradeoffs among metrics.
Our contributions are as follows.
\begin{itemize}
    \item To the best of our knowledge, no existing work considers multi-dimensional diversity for training level generators. Our novel approach marks a significant contribution to expanding horizons of diversity in generating levels, addressing Challenge 1.

    \item 
    We formalise the training of generative models as a multi-objective learning problem, where each metric for evaluating generated levels is treated as a distinct objective. Our proposed multi-objective evolutionary learning framework optimises multi-dimensional diversity metrics simultaneously, without the need for assigning weights among these metrics to convert them into one, thereby dealing with Challenge 2.

    \item 
    Our framework can generate a diverse set of generators, as Pareto front, in a single run, and effectively captures the tradeoffs among the considered metrics, accommodating diverse preferences of users with varying priorities, thus resolving Challenge 3.
    
\end{itemize}

Extensive experimentation on the commonly used \emph{Super Mario Bros.} (SMB) benchmark validates our approach. Results confirm that our algorithm, based on this framework, optimises three objectives more effectively. These objectives include a playability metric, a player-centered diversity metric, and a content-based diversity metric. These experimental outcomes underscore our framework's capability in enhancing multi-dimensional diversity for game levels.

This paper is organised as follows. Section \ref{sec:bak} introduces the background of multi-objective optimisation in generating video game levels, diversity metrics, and methods for enhancing diversity in video game levels. Section \ref{sec:frame} outlines our proposed framework and the implemented algorithm. Experimental results are presented in Section \ref{sec:exp}. Finally, Section \ref{sec:conclusion} concludes the paper and discusses future work.

\section{Background}\label{sec:bak}
This section introduces the definitions of multi-objective optimisation and its application in generating video game levels. Next, the diversity metrics within video game levels are reviewed. Following this, we summarise existing methods for enhancing diversity in video game levels.

\subsection{Multi-objective Optimisation and Its Application in Video Game Level Generation}

Many real-world problems, including those in generating video game levels, require simultaneously optimising multiple conflicting objectives. Such problems can be characterised as multi-objective problems (MOPs), formulated as follows:
\begin{equation}
\begin{aligned}
   &minimise\ F(\bm{x}) = \{ f_1(\bm{x}), \dots, f_m(\bm{x}) \}, \\
   &subject \ to: \bm{x} \in \Omega,
\end{aligned}
\label{equ:MOP}
\end{equation}
where $\Omega$ represents the decision space and $\bm{x} \in \Omega$ are the decision variables. $F(\bm{x})$ contains $m$ objectives $f_1(\bm{x}),\dots,f_m(\bm{x})$, mapping $\bm{x}$ to $\Phi^m$, where $\Phi^m$ denotes the objective space, i.e., $F$: $\Omega \rightarrow \Phi^m$. The objectives within $F(\bm{x})$ typically conflict, implying that optimising one objective may lead to the degradation of at least one other objective. 

The inherent conflicting nature of objectives in multi-objective optimisation suggests the existence of a set of tradeoffs~\cite{li2019quality}, known as Pareto-optimal solutions (PS) within the decision space $\Omega$. The Pareto front (PF) represents the corresponding set in the objective space of PS. A variety of multi-objective evolutionary algorithms (MOEAs)~\cite{ZHOU201132, 10.1145/2792984} have been proposed to address MOPs. A solution set that exhibits both high convergence and diversity is highly desirable for decision-makers. Such a set not only provides high-quality solutions but also assists decision-makers in understanding the underlying problem by showcasing various tradeoffs. 

In game level generation, conflicts may arise not only within multiple level diversity metrics but also between diversity and other metrics (e.g., playability) as well. From an MOP viewpoint, if designers aim to optimise playability and multiple level diversity metrics, there should be a set of tradeoffs representing the PF among these objectives. In the context of this work, for a finite solution set, two critical quality aspects are convergence and population diversity~\cite{li2019quality}. Convergence refers to the closeness of the solution set to the PF, while population diversity focuses on the degree of dissimilarity within the solution set, considering their respective objective values. Solutions covering various degrees in terms of objectives, e.g., content-based and player-centered diversity metrics, indicate good diversity. Notably, the concept of ``diversity'' in MOP is distinct from that in PCG. In MOP, diversity focuses on the variation in tradeoffs among different objectives. In PCG, the diversity emphasises the creativity and innovation in game content generation, such as measuring diversity on playtrace.

Several works have applied multi-objective optimisation to PCG. Lara-Cabrera \textit{et al.} consider two conflicting objectives with NSGA-II when generating maps for real-time strategy games~\cite{lara2014balance}, while Ruela \textit{et al.} compare four classic multi-objective evolutionary algorithms in generating unpredictable and diverse levels~\cite{ruela2020multi}. Zhang \textit{et al.}~\cite{zhang2024interpreting} demonstrate the capability of Two\_Arch2~\cite{wang2014two_arch2} in generating \textit{Sokoban} levels considering two conflicting objectives, spatial diversity and emptiness diversity, by searching in the tile-based level space. Those works, however, focus on directly searching levels themselves, differing from our approach, which concentrates on training generators. Khalifa and Togelius introduce a framework named \textit{Marahel} which represents constructive level generators through domain-specific grammar, and employs NSGA-II~\cite{deb2002fast} to evolve controllable level generators~\cite{khalifa2020multi}. But these approaches often rely on expert knowledge to construct grammar. 

The approaches utilised to train generators, such as GANs~\cite {10.1145/3422622} and VAEs~\cite{DBLP:journals/corr/KingmaW13}, have yielded significant advancements in game level generation~\cite{volz2018evolving,snodgrass2020multi}. These approaches offer greater flexibility and adaptability and overcome challenges such as domain-specific grammar. Nonetheless, these techniques often yield a single tradeoff among considered objectives, such as diversity, thereby constraining their real-world applicability. For example, given a set of considered diversity objectives, a decision-maker may prefer a specific tradeoff among these objectives. These methods can tailor a generator to accommodate a suitable tradeoff by carefully calibrating weights among the objectives within a single run. When a decision-maker requires a new tradeoff for a different scenario, they must identify another set of weights to accommodate the new tradeoff, resulting in significant additional costs. 

Therefore, a method providing a set of generators representing diverse tradeoffs within one run is highly desirable for decision-makers. A such method not only reduces the burden on decision-makers by eliminating the need for determining alternative weight sets but also enhances the adaptability of generators in addressing varying scenarios.

\subsection{Measuring Diversity in Video Game Levels}

Diversity in video game levels can be quantitatively assessed through a variety of metrics~\cite{liy2024measuring}, as evidenced by numerous studies in the field, e.g.,~\cite{preuss2014searching, wang2022fun,beukman2022procedural,kutzias2023recent,withington2023right}. However, there is a lack of a universally accepted diversity metric~\cite{Jordanous2011Evaluating,liy2024measuring}. Generally, diversity metrics in game levels can be divided into two main categories~\cite{Shaker2016Evaluating}: content-based and player-centered metrics. Content-based metrics specifically assess the diversity on the variety, such as tiles, elements, patterns or maps, within game levels; in contrast, player-centered metrics assess diversity on player behaviours, which can be either those of humans or modelled by agents.

Existing works in game level generation commonly utilise distance metrics to quantify the dissimilarity between two  levels~\cite{beukman2022procedural,zakaria2023procedural,earle2021learning,marino2015empirical}. Formally, diversity metric based on distance metrics is expressed as $\expect[\delta(x, x')]$, where $\delta(\cdot, \cdot)$ represents a distance metric, and $x$ and $x'$ are two levels. In the context of content-based metrics, Earle \textit{et al.} utilise the Hamming distance which quantifies the numbers of different tiles between two levels~\cite{earle2021learning}. Lucas and Volz propose tile-pattern Kullback-Leibler divergence as a measure to evaluate the differences in tile patterns of two levels~\cite{lucas2019tile}. While Kullback-Leibler divergence is sensitive to uncommon tile patterns in the two levels, Wang \textit{et al.}~\cite{wang2022fun} employ Jensen-Shannon divergence (JS-divergence) as an alternative. To assess player-centered diversity, Beukman \textit{et al.}~\cite{beukman2022procedural} utilise Levenshtein distance while Wang \textit{et al.}~\cite{wang2022fun} apply dynamic time warping (DTW)~\cite{berndt1994using} as distance measure.

These diversity metrics, originally designed to quantify diversity among levels, can be effectively adapted to evaluate the generator diversity. An intuitive approach is to calculate the diversity of a number of levels generated by a given generator~\cite{gravina2019procedural}. This method offers a comprehensive insight into the generator's capability of producing varied levels, regardless of content-based metrics or player-centered metrics.

\subsection{Generating Diverse Video Game Levels}

A traditional way of generating diverse levels is novelty search \cite{lehman2011abandoning}, which uses an objective function to evaluate the novelty of newly generated levels relative to the previous generations \cite{preuss2014searching,liapis2013enhancements,sudhakaran2024mariogpt}. The novelty function can be considered as an assessment of the contribution from a new level to the diversity of the whole level set. By optimising this novelty function, novel levels can be gradually added to an archive, resulting in a set of diverse levels. Quality diversity search \cite{gravina2019procedural}, increasingly used in recent works \cite{fontaine2021illuminating,medina2023evolving},  maintains an archive of generated levels partitioned into cells representing different attributes, which indicates a diverse range of content. High-quality levels are assigned to specific cells, resulting in a diverse set of levels with various attributes.

The aforementioned methods directly generate diverse levels gradually by adding new and either distant or behaviourally different levels to a game-level set. Unlike this kind of direct method, there has been a recent trend of training machine learning models as level generators serving a variety of applications, including experience-driven PCG \cite{Sarkar2024Procedural}, interactive PCG \cite{lai2022mixed} and online PCG \cite{shu2021experience}. Those generators can improve the efficiency of generating content, making it possible to generate large-scale game content like 3D worlds \cite{AwiSch2023worldgan}.

Several studies consider enhancing the diversity of levels generated by a learning-based generator. Torrado \textit{et al.}~\cite{torrado2020bootstrapping} suggest adding levels generated by the generator back into the training set to enrich the training set, promoting generator diversity. Bontrager and Togelius~\cite{bontrager2021learning} introduce a diversity loss in training a level generator to maximise the expected distance between two generated levels. Beukman \textit{et al.}~\cite{beukman2022procedural} combine neuroevolution and novelty search to train level generators and employ an intra-generator novelty fitness function to enable the generation of diverse levels. The work of~\cite{wang2022fun} considers two kinds of diversity and applies a weighted sum approach to convert them into one objective to update a level generator.

As far as we know, there is no prior research that addresses multi-dimensional diversity in training multiple generative models for game levels, where each model represents a specific tradeoff within the multi-dimensional diversity. Our work aims to fill this gap, demonstrating how multi-dimensional generator diversity can significantly benefit from a multi-objective approach.

\section{Multi-objective Evolutionary Learning Framework for Generating Multi-dimensional Diverse Levels}\label{sec:frame}
In this section, we propose a multi-objective evolutionary learning framework for training generative models to enhance multi-dimensional diversity in game level generation. Then, based on this framework, an algorithm for training SMB generators is implemented as a case study.

\subsection{Framework Overview}

Our proposed framework, outlined in Algorithm \ref{algo:framework}, aims to evolve a population of generative models for generating content, not necessarily levels. Each individual in the population represents a generative model. It aims to guide the population towards better overall performance tradeoffs among playability and multiple diversity metrics from an MOP perspective.

The framework's input includes a game content dataset $\mathcal{C}$, a set of generative model evaluation metrics $\mathcal{F}$, the size of model population $\lambda$ and a multi-objective optimiser $\pi$. The optimiser $\pi$ primarily incorporates strategies for mating selection, reproduction and survival selection. First, we randomly initialise $\lambda$ generative models as $\mathcal{G} = \{G_1,\dots,G_\lambda\}$. The models are evaluated by all metrics in $\mathcal{F}$. Within the main loop, the mating selection strategy of $\pi$ selects $k$ promising models as parent models $\mathcal{P}$ (line \ref{line:parentselection} in Algorithm \ref{algo:framework}). Subsequently, $\mu$ new models, $\mathcal{G}'$, are generated (line \ref{line:generateoff} in Algorithm \ref{algo:framework}) through reproduction strategy of $\pi$, inheriting information from $\mathcal{P}$. The reproduction strategy modifies the parameters of parent models to create new models as offspring. Then, $\lambda$ models are selected from $\mathcal{G} \cup \mathcal{G}'$ through the survival selection of $\pi$ as the population of the next generation (line \ref{line:envselection} in Algorithm~\ref{algo:framework}). This process repeats until a termination criterion is met.

Employing MOEAs~\cite{ZHOU201132, 10.1145/2792984} as the multi-objective optimiser $\pi$ is ideal for evolving a set of learning models focusing on convergence and diversity, e.g., \cite{Mitigating_unfairness_2023,fairerML_2021}. In the implementation of our framework through $\pi$, the critical steps involve evaluation models using multiple metrics, and generating new models based on these evaluations. The output models produced by this process can be further curated and refined by human decision-makers, aligning with various requirements and objectives. Our framework is general and flexible, comprising modules such as evaluation metrics and a multi-objective optimiser. While the selection of these modules depends on the specific requirements of the problem at hand, the algorithm is designed to be easily adaptable, requiring minimal adjustments to be applied to other problems.

\begin{algorithm}[htbp]\color{black}
\caption{\label{algo:framework}Multi-objective learning framework for training level generators.}
\begin{algorithmic}[1]
\Require Game content dataset $\mathcal{C}$, set of generative model evaluation metrics $\mathcal{F}$, size of generative model set $\lambda$, multi-objective optimiser $\pi$
\Ensure A model set $\mathcal{G} = \{G_1,\dots,G_\lambda\}$
\State Randomly initial generative models $\mathcal{G} = \{G_1,\dots,G_\lambda\}$
\State {$\{ \mathcal{F}(G_1), \dots,  \mathcal{F}(G_\lambda)\} \leftarrow$ Evaluate each model of ${\mathcal{G}}$ with metrics $\mathcal{F}$ \label{line:eval1}}

\While{terminal conditions are not fulfilled}
    \State $\mathcal{P} \leftarrow$ Select $k$ models from $G_1,\dots,G_\lambda$ with best \NoNumber{$\mathcal{F}(G_1), \dots,  \mathcal{F}(G_\lambda)$ according to $\pi$} \label{line:parentselection}
    \State {$\mathcal{G}' \leftarrow$ Generate $\mu$ new models $G_1',\dots, G_\mu'$ from $\mathcal{P}$ \NoNumber{according to $\pi$ on $\mathcal{C}$} }\label{line:generateoff}
    \State {$\{ \mathcal{F}(G'_1), \dots,  \mathcal{F}(G'_\mu)\} \leftarrow$ Evaluate each model of ${\mathcal{G}'}$ \NoNumber{with metrics $\mathcal{F}$}\label{line:eval2}}
    \State{$\{(G_1,\mathcal{F}(G_1)),\dots, (G_\lambda,\mathcal{F}(G_\lambda))\} \leftarrow$ Select $\lambda$ models \NoNumber{from $\mathcal{G} \bigcup \mathcal{G}'$} by $\pi$ based on $\{\mathcal{F}(G_1), \dots,  \mathcal{F}(G_\lambda)\}$} and \NoNumber{$\{\mathcal{F}(G'_1), \dots,  \mathcal{F}(G'_\mu)\}$, then update $G_1,\dots,G_\lambda$ and \NoNumber{their corresponding evaluation values}} \label{line:envselection}
\EndWhile
\end{algorithmic}
\end{algorithm}

\subsection{Proposed Algorithm based on Our Framework}\label{sec:algorithm}

The selection of the model set, evaluation metrics, and multi-objective optimisation algorithm in our proposed framework can vary depending on the specific requirements of the tasks and preferences. In our work, generating levels for SMB is taken as a case study. SMB level generation is a commonly used benchmark problem for researching procedural level generation~\cite{togelius2011search,summerville2018procedural,liu2021deep,hu2024games}. 

We have developed an algorithm derived from our framework, utilising key elements as follows with the detailed processes outlined in Algorithms \ref{algo:GAN_framework}, \ref{algo:variation} and \ref{algo:Warm}. In summary, Algorithm~\ref{algo:GAN_framework} is the instantiation's backbone built upon our framework, as shown in Algorithm~\ref{algo:framework}, where a GAN is used as the generator. Algorithms~\ref{algo:variation} and~\ref{algo:Warm} are the detailed modules in implementing the instantiation. The former describes the variation strategy, which creates offspring generators based on parent generators, while the latter details the warm start process for training a set of generative models and a discriminator.

\begin{algorithm}[htbp]
\caption{\label{algo:GAN_framework}Multi-objective learning framework for generating game levels using GAN.}
\begin{algorithmic}[1]
\Require Game level dataset $\mathcal{C}$, set of generative model evaluation metrics $\mathcal{F}$, size of generative model set $\lambda$, multi-objective optimiser $\pi$, warm start generation $T_w$, maximum evolution generation $T$, iteration of training discriminator $T_d$, batch size $b$, Adam optimiser $Adam$
\Ensure A model set $\mathcal{G} = \{G_1,\dots,G_\lambda\}$
\State Initial current generation $t = 1$
\State $\{G_1,\dots, G_\lambda\}$ and $D$ $\leftarrow$ Perform warm start (cf. Algorithm \ref{algo:Warm}) to obtain a generator set and a discriminator 
\State {$\{ \mathcal{F}(G_1), \dots,  \mathcal{F}(G_\lambda)\} \leftarrow$ Evaluate each model of ${\mathcal{G}}$ with metrics $\mathcal{F}$}
\While{$t < T$}
    \For{$e \leftarrow 1, \dots, T_d$}
    \Statex \qquad $//$ \textit{Update the discriminator model}
        \For{batch $c \subset \mathcal{C}$}
            \State {$z \leftarrow$ Generate a noise batch from a Gaussian \NoNumber{distribution}}
            \State {$g_D \leftarrow \nabla_D [-\frac{1}{b} \sum_{i=1}^{b} \min(0, -1 + D(c_i))-$}
            \Statex {\qquad \qquad \qquad$\frac{1}{b}\sum_{j=1}^{n}\sum_{s=1}^{b/n} \min(0, -1 - D(G_j(z_s))) ]$}
            \State {Update the parameters of $D$ based on $g_D$ \NoNumber{using $Adam$}}
        \EndFor
    \EndFor
    \State {$\mathcal{G}' \leftarrow$ Perform variation strategy (cf. Algorithm \ref{algo:variation}) using $\mathcal{G}$, $D$ and \NoNumber{$Adam$ to generate new generators}} \label{line:variation}
    \State {$\{ \mathcal{F}(G'_1), \dots,  \mathcal{F}(G'_\mu)\} \leftarrow$ Evaluate each model of ${\mathcal{G}'}$ \NoNumber{with metrics $\mathcal{F}$}}\label{line:grtor_eval}
    \State{$\{(G_1,\mathcal{F}(G_1)),\dots, (G_\lambda,\mathcal{F}(G_\lambda))\} \leftarrow$ 
    Select $\lambda$ models \NoNumber{from $\mathcal{G} \bigcup \mathcal{G}'$} by $\pi$ based on $\{\mathcal{F}(G_1), \dots,  \mathcal{F}(G_\lambda)\}$} and \NoNumber{$\{\mathcal{F}(G'_1), \dots,  \mathcal{F}(G'_\mu)\}$, then update $G_1,\dots,G_\lambda$ and \NoNumber{their corresponding metric evaluations}} \label{line:envselection2}
    \State $t = t + 1$
\EndWhile

\end{algorithmic}
\end{algorithm}

\begin{algorithm}[htbp]
\caption{\label{algo:variation}Variation strategy. }
\begin{algorithmic}[1]
\Require A set of generators $\mathcal{G} = \{G_1,\dots,G_\lambda\}$, a discriminator $D$, Adam optimiser $Adam$, batch size $b$, iteration of training generator $T_g$
\Ensure Newly generated generators $\mathcal{G}'$
\State {$\mathcal{G}'=\emptyset$}
\For{$j \leftarrow 1, \dots, \lambda$}
\Statex \quad $//$ \textit{Perform the minmax mutation strategy}
\State{$G = copy(G_j)$}
\For{$e \leftarrow 1, \dots, T_g$}
    \State {$z \leftarrow$ Generate a noise batch from a Gaussian \NoNumber{distribution} \label{line:mu_generate}}
    \State {$g_{G} \leftarrow \nabla_{G} \left[ \frac{1}{b}\sum_{i=1}^{b} -D(G(z_i))\right] $} \label{line:minmax}
    \State {$G \leftarrow$ Update the parameters of $G$ based on $g_{G}$\NoNumber{ using $Adam$} \label{line:mu_adam}}
\EndFor
\State {$\mathcal{G}'= \mathcal{G}' \bigcup \{ G\}$}
\Statex \quad $//$ \textit{Perform the Least-squares mutation strategy}
\State{$G = copy(G_j)$}
\For{$e \leftarrow 1, \dots, T_g$}
    \State {$z \leftarrow$ Generate a noise batch from a Gaussian \NoNumber{distribution}\label{line:mu2_noise}}
    \State {$g_{G} \leftarrow \nabla_{G} \left[ \frac{1}{b}\sum_{i=1}^{b} (D(G(z_i))-1)^2   \right] $}\label{line:Least}
    \State {$G \leftarrow$ Update the parameters of $G$ based on $g_{G}$\NoNumber{ using $Adam$}\label{line:mu2_adam}}
\EndFor
\State {$\mathcal{G}'= \mathcal{G}' \bigcup \{ G\}$}
\EndFor

\end{algorithmic}
\end{algorithm}

\begin{algorithm}[htbp]
\caption{\label{algo:Warm}Warm start for training a set of GANs.}
\begin{algorithmic}[1]
\Require Game level dataset $\mathcal{C}$, size of generative model set $\lambda$, batch size $b$, training epoch $T_w$, iteration of training generator $T_g$, Adam optimiser $Adam$
\Ensure A model set $\mathcal{G}$ and a discriminator $D$
\State {Randomly initial generator models $\mathcal{G} = \{G_1,\dots,G_\lambda\}$\label{line:warm_init_G}}
\State {Randomly initial a discriminator model $D$\label{line:warm_init_D}}

\For{$e \leftarrow 1, \dots, T_w$}
    \For{batch $c \subset \mathcal{C}$}
        \Statex \qquad $//$ \textit{Update the discriminator model }
        \State {$z \leftarrow$ Generate a noise batch from a Gaussian \NoNumber{distribution}\label{line:warm_noise}} \label{line:d1}
        \State {$g_D \leftarrow \nabla_D [-\frac{1}{b} \sum_{i=1}^{b} \min(0, -1 + D(c_i))-$ }
        \Statex {\qquad \qquad \qquad$\frac{1}{b}\sum_{j=1}^{n}\sum_{s=1}^{b/n} \min(0, -1 - D(G_j(z_s))) ]$ \label{line:warm_D_grad}}
        \State {Update the parameters of $D$ based on $g_D$ \NoNumber{using $Adam$}} \label{line:d2}
        \Statex \qquad  $//$ \textit{Update each generator model\label{line:warm_D_adam}}

        \For{$j \leftarrow 1, \dots, \lambda$}
            \State {$z \leftarrow$ Generate a noise batch from a Gaussian \NoNumber{distribution}\label{line:G_1}} 
            \State {$g_{G_j} \leftarrow \nabla_{G_j} \left[ \frac{1}{b}\sum_{i=1}^{b} -D(G_j(z_i))\right] $ \label{line:warm_G_grad}}
            \State {Update the parameters of $G_j$ based on $g_{G_j}$ \NoNumber{using $Adam$}} \label{line:G_2}
        \EndFor\label{line:warm_start_ed}
  
    \EndFor
\EndFor
   
\end{algorithmic}
\end{algorithm}

\subsubsection{Model set} 
A variety of generative models can be applied in this context. In this work, self-attention GAN (SAGAN) architecture~\cite{zhang2019self} is employed as individuals, as it has shown promising results in game level generation~\cite{wang2024negatively}. In our implementation, each individual maintains its own generator, while sharing a single discriminator among them, following previous studies in evolving a set of GANs~\cite{wang2019evolutionary,MOGAN2020}.

\subsubsection{Evaluation metrics} 
In our work, three representative metrics are considered: a playability metric ($P$), a player-centered diversity metric, $PD$, and a content-based diversity metric, $CD$. $PD$ and $CD$ represent two distinct dimensions of diversity metrics for game levels, and are complementary to each other. The assessment of $PD$ is based on behaviours exhibited by a gameplaying agent. Let $c_i$ denote the $i$th level sample. Those metrics are formulated as follows.

$P$ estimates the probability of generating a playable level for a generator. An SMB level is considered \textit{playable} if a game-playing agent can complete it; otherwise, it is regarded as unplayable~\cite{shaker2016procedural,shu2021experience,fontaine2021illuminating,khalifa2020pcgrl,volz2018evolving}, formulated as 
\begin{equation}\label{eq:PP}
    \mathrm{P} = \frac{1}{n} \sum_{i=1}^{n} \mathbbm{1} [c_i \text{ is completed by the agent}].
\end{equation}

$PD$, formulated in Eq.~\eqref{eq:PD}, employs the DTW~\cite{berndt1994using} distance averaged across levels to measure the diversity between playtraces on generated levels~\cite{wang2022fun}:
\begin{equation}\label{eq:PD}
    \mathrm{PD} = \frac{2}{n(n-1)} \sum_{i=1}^{n} \sum_{j=1}^{n} \mathrm{DTW}(\tau(c_i), \tau(c_j)),
\end{equation}
where $\tau(c)$ is the agent's playtrace in level $c$. Higher $PD$ values indicate higher player-centered diversity.

$CD$, formulated in Eq.~\eqref{eq:CD} uses tile-pattern Jensen–Shannon divergence (TPJS) \cite{wang2022fun} to estimate the dissimilarity in distributions of tile patterns between levels. The average TPJS is used as a content-based diversity assessment: 
\begin{equation}\label{eq:CD}
    \mathrm{CD} = \frac{2}{n(n-1)} \sum_{i=1}^{n} \sum_{j=1}^{n} \mathrm{TPJS}(P(c_i), P(c_j)),
\end{equation}
where $P(c)$ is the tile-pattern frequency distribution of level $c$ \cite{lucas2019tile}. Higher $CD$ values signify higher content-based diversity.

To align with the minimisation objective of Eq.~\eqref{equ:MOP}, we apply simple linear transformations to $P$, $PD$, and $CD$. These transformed metrics, denoted as $f_{P}$, $f_{PD}$ and $f_{CD}$, are restructured to be minimised. Specifically, we define $f_{P} = 1 - \mathrm{P}$, $f_{PD} = \frac{200 - \mathrm{PD}}{100}$ and $f_{CD} = 1 - \mathrm{CD}$. Such linear transformations of objectives just scale the objective values to the same range without affecting optimisation results. Note that evaluating diversity metrics of levels may require simulating game-playing agents in generated levels, which can be computationally time-consuming. Utilising surrogate models to approximate metric values~\cite{8456559} could be an efficient way to reduce the cost, serving as a potential future work.

\subsubsection{Multi-objective optimiser}
In Algorithm \ref{algo:GAN_framework}, built upon our framework (Algorithm \ref{algo:framework}), the survival selection component of $\pi$ (line \ref{line:envselection2}) is implemented using SDE$^+$-MOEA~\cite{SDEPMOEA_2023} for its efficiency and hyperparameter-free. According to the study~\cite{SDEPMOEA_2023}, the computational complexity of our algorithm is $O(\lambda^2 \log(\lambda))$, where $\lambda$ is the population size.

For reproduction (line \ref{line:variation} in Algorithm \ref{algo:GAN_framework}), minmax and least-squares, two mutation strategies commonly applied to GANs~\cite{wang2019evolutionary,MOGAN2020} are incorporated. Given the use of SAGAN instead of vanilla GAN, these two strategies are modified, as detailed in Algorithm \ref{algo:variation}. Two loss functions (lines \ref{line:minmax} and \ref{line:Least}) are utilised to update the parameters of each parent generator using a noise batch sampled from a Gaussian distribution.

Specifically, Algorithm \ref{algo:variation} starts with an empty set of generators, $\mathcal{G}'$, and iteratively adds new generators by applying two mutation strategies to each generator $G_j$ in the set $\mathcal{G}$. For each generator, it first creates a copy and then applies the minmax mutation strategy by generating noise batches (line~\ref{line:mu_generate}), computing gradients (line~\ref{line:minmax}), and updating the generator's parameters using the Adam optimiser (line~\ref{line:mu_adam}) over $T_g$ iterations. The updated generator is then added to $\mathcal{G}'$. Next, another copy of $G_j$ undergoes the least-squares mutation strategy, following similar steps (lines~\ref{line:mu2_noise}--\ref{line:mu2_adam}). This updated generator is also added to $\mathcal{G}'$. By the end of the process, $\mathcal{G}'$ contains new generators obtained through these two mutation strategies.

\subsubsection{Process of our proposed algorithm}
The main process of our proposed method is outlined in Algorithm \ref{algo:GAN_framework}, incorporating the core elements discussed above. The process begins with a warm start method, outlined in Algorithm \ref{algo:Warm}, to establish a set of well-pre-trained SAGANs. Specifically, the process starts with a randomly initialised set of generators $\mathcal{G} = \{G_1, \dots, G_\lambda\}$ (line~\ref{line:warm_init_G}) and a discriminator $D$ (line~\ref{line:warm_init_D}). Over $T_w$ epochs, for each batch of game levels $c$ from dataset $\mathcal{C}$, the gradients $g_D$ for updating the discriminator $D$ are calculated based on real $c_i$ and generated data $G_j(z_s)$ (line~\ref{line:warm_D_grad}), and adjust $D$'s parameters using the Adam optimiser (line~\ref{line:d2}). Each generator $G_j$ is then updated individually. For each generator, noise $z$ is generated (line~\ref{line:G_1}) and then gradients for updating each generator $g_{G_j}$ are computed (line~\ref{line:warm_G_grad}). $G_j$'s parameters are updated using the Adam optimiser (line~\ref{line:G_2}).

After the warm start phase, the generators are evaluated using the metrics $f_{P}$, $f_{PD}$ and $f_{CD}$, formulated in Eq.~\eqref{eq:PP}, Eq.~\eqref{eq:PD}, and Eq.~\eqref{eq:CD}, respectively. The discriminator is then updated, similar to the warm start phase. Subsequently, the process of variation (Algorithm \ref{algo:variation}), offspring generator evaluation and generator selection (implemented by SDE$^+$-MOEA) are executed, as shown in the lines \ref{line:variation}, \ref{line:grtor_eval} and \ref{line:envselection2} in Algorithm~\ref{algo:GAN_framework}. Finally, Algorithm~\ref{algo:GAN_framework} outputs a Pareto set of generators.

\section{Experimental Studies}\label{sec:exp}
This section delves into the experimental evaluation of our framework. We begin by detailing the comparison methods and then present the benchmark and parameter settings used in our study. Subsequently, the effectiveness of our framework is verified from two perspectives, including enhancing multi-dimensional diversity and providing a set of diverse generators.

\subsection{Compared Methods}
To comprehensively assess the effectiveness of our framework in addressing the multi-dimensional diversity of game-level generation, two representative diversity metrics are used. However, it's worth highlighting that our findings can be generalised to scenarios involving more than two diversity metrics. The three metrics considered are a playability metric ($f_{P}$), a player-centered diversity metric ($f_{PD}$), and a content-based diversity metric ($f_{CD}$), described in Section \ref{sec:algorithm}. Overall, the following four cases are included in this comparative analysis.
\begin{itemize}
    \item $A_{P+PD+CD}$ is a tri-objective case considering all the three metrics, i.e., $f_{P}$, $f_{PD}$ and $f_{CD}$. 
    \item $A_{P+PD}$ is bi-objective case considering $f_{P}$ and $f_{PD}$.
    \item $A_{P+CD}$ is bi-objective case considering $f_{P}$ and $f_{CD}$.
    \item $A_{P}$ only considers $f_{P}$, i.e., the playability.
\end{itemize}

\subsection{Benchmark}
The widely used Mario AI framework \cite{karakovskiy2012mario}, an open-source implementation of Nintendo's SMB is used. While SMB represents a specific genre and may not capture the full spectrum of game complexities, it provides a valuable testbed for a series of games and has been widely used in validating level generation methods~\cite{togelius2013procedural,sudhakaran2024mariogpt,fontaine2021illuminating,shu2021experience,awiszus2020toad,volz2018evolving}.
\begin{table}[htbp]
    \centering
    \caption{Tile types and corresponding images used in this work. Adjacent pipe tiles will be parsed into pipe objects and be assigned the proper image.}
    \input{tables/representation}
    \label{tab:representation}
\end{table}

Within the Mario AI framework, SMB levels are represented by 2D token arrays, with each token representing a different tile type. In this study, we consider $21$ distinct tile types. Details of these tile types, along with their corresponding images, are presented in Table \ref{tab:representation}. Furthermore, the framework features a game engine for search-based agents and includes several built-in agents to facilitate gameplay simulation. For the purpose of our experiments, we have selected the agent, used in \cite{zhang2022generating}, as our simulation agent.

\subsection{Parameter Setting}
Table \ref{tab:parameter} summarises the parameters of our algorithm used in the experiments. The specific hyperparameters for updating the generators and discriminator follow the configurations in~\cite{wang2024negatively}.

\begin{table}[htbp]
  \centering
  \caption{Parameter setting}
  \setlength\tabcolsep{0.8pt} {
    \begin{tabular}{cccc}
    \toprule
    \multicolumn{2}{c}{Different aspects} & \multicolumn{2}{p{21em}}{\thead{Detailed settings}} \\
    \midrule
    \multicolumn{2}{c}{Evolving GANs} & \multicolumn{2}{p{21em}}{\thead{ Number of generators $\lambda$: 30\\Epoch of warm start $T_w$: 100\\ Maximum generations $T$: 100}} \\
    \midrule
    \multirow{4}[4]{*}{\thead{Training\\ a GAN}} & Generator & \multicolumn{1}{p{12.125em}}{\thead{Learning rate: 1e-4\\Weight decay: 0\\Size of noise: 128\\ Training iteration $T_g$: 1}} & \multicolumn{1}{c}{\multirow{4}[4]{*}{\thead{Batch size $b$: 32\\\thead{Adam: beta1=0,\\ beta2=0.9}}}} \\
\cmidrule{2-3}          & Discriminator & \multicolumn{1}{p{12.125em}}{\thead{Learning rate: 4e-4\\Weight decay: 5e-4\\Training iteration $T_d$: 1}} &  \\
    \midrule
    \multicolumn{2}{c}{\thead{Metric\\evaluation}} & \multicolumn{2}{c}{\thead{Number of level samples $n$: 30\\ Gameplaying agent: \textit{Collector}}} \\
    \bottomrule
    \end{tabular}%
    }
  \label{tab:parameter}%
\end{table}%

\subsection{Performance Measure}

Two widely used indicators in the multi-objective optimisation literature~\cite{li2019quality}, hypervolume (HV)~\cite{Shang2021Survey} and coverage over the Pareto front (CPF)~\cite{Tian2019Diversity}, are employed to evaluate the performance of the generator sets obtained through the comparison methods. Larger HV or CPF values indicate better performance. HV~\cite{Shang2021Survey} is used to evaluate the overall performance of the obtained generator sets, considering both convergence and diversity, which is applied to validate the effectiveness of enhancing the multi-dimensional diversity of generator sets. CPF~\cite{Tian2019Diversity}, on the other hand, focuses more on assessing diversity of generator sets in terms of playability and diversity metrics~\cite{li2019quality, Tian2019Diversity}. Therefore, CPF is used to quantify the diversity of generator sets concerning $f_{P}$, $f_{PD}$ and $f_{CD}$.

Due to the unknown true PF, the computation of HV and CPF involves aggregating all non-dominated generators discovered across all experimental trials that consider the same objectives. These solutions collectively form a pseudo PF. Specifically, the pseudo PF considering $f_{P}$, $f_{PD}$ and $f_{CD}$, incorporates all generators from every generation across 30 trials of the comparison methods. The objective values of the generator sets are then normalised to the closed intervals $[0,1] \times \cdots \times [0,1]$ based on this pseudo PF. The nadir point in HV is set to $\{1.1,\dots, 1.1\}$. Then, the average HV and CPF values for each generation are recorded throughout each experiment, encompassing all 30 trials. This approach ensures that the HV and CPF calculation involves the contributions of $f_{P}$, $f_{PD}$, and $f_{CD}$, thereby ensuring the comparability of HVs and CPFs among different algorithms with varying optimisation objectives.

\subsection{Effectiveness of Enhancing Multi-dimensional Diversity}

To comprehensively evaluate the ability of our framework to enhance multi-dimensional diversity in game-level generation, we approach this validation from three perspectives. First, the overall performance of generator sets, considering both convergence and diversity, is assessed through HV. Specifically, we visualise the evolution process through HV and assess the statistical HV performance of the comparison methods,  indicating the advantages of our framework for enhancing multi-dimensional diversity. Second, the distribution of tradeoffs within one arbitrary trial obtained by comparison methods is visualised to explain the reasons for the observed advantages. Third, we record the distribution of the knee point generators across all trials to gain statistical insights into the behaviours of the comparison methods concerning the three objectives. The knee point~\cite{6975108,das1999characterizing} of a PF represents a balance where no single objective is excessively sacrificed. In other words, the knee point can not only indicate the extent of the ``central'' compromised tradeoff but also provide a tradeoff to decision-makers who do not particularly prioritise one objective over others. The detailed experimental implementations and results of the three aforementioned perspectives are introduced below.

\begin{figure}[hb]
    \centering
    \includegraphics[scale=0.36]{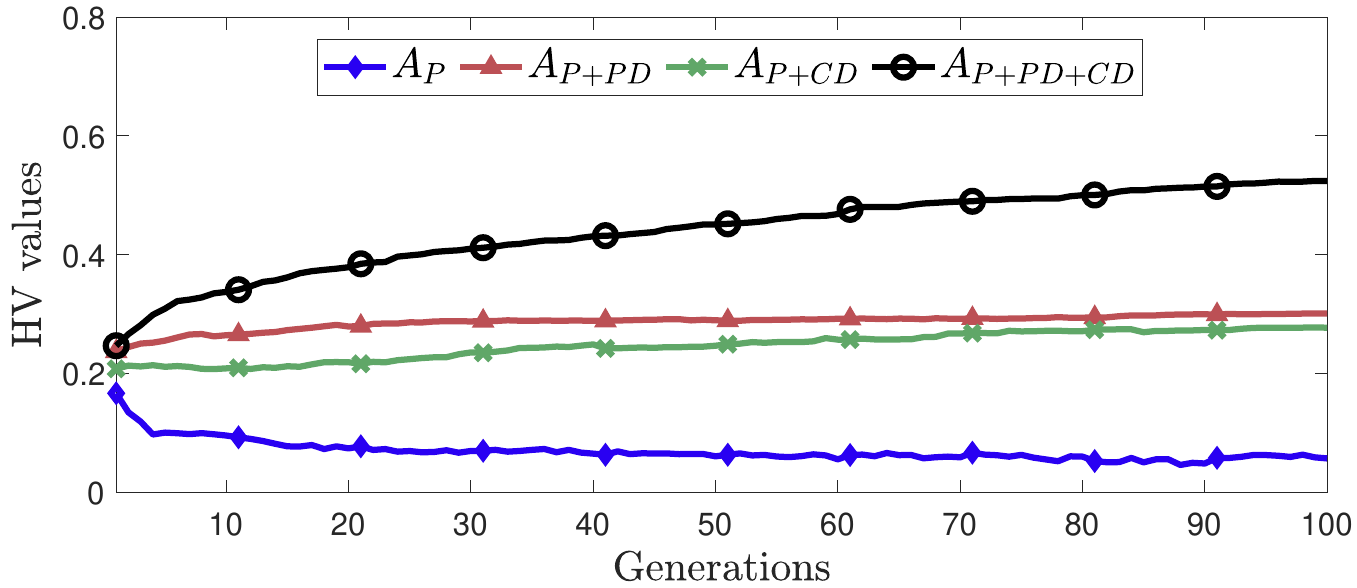}
    \caption{Averaged HV values over 30 trials. }
    \label{fig:HV}
\end{figure}

\paragraph{HV performance}
Fig. \ref{fig:HV} illustrates the convergence curves of HV values across 30 independently repeated trials of $A_{P+PD+CD}$, $A_{P+PD}$, $A_{P+CD}$ and $A_{P}$, quantifying the optimisation process. Here, $f_{PD}$ and $f_{CD}$ represent two dimensions of diversity, which $f_{PD}$ focuses on the diversity of playtraces, and $f_{CD}$ emphasises level diversity. Each of $A_{P+PD+CD}$, $A_{P+PD}$, $A_{P+CD}$ and $A_{P}$ directly optimises different combinations of the four metrics. First, Fig. \ref{fig:HV} demonstrates that the HV curves of methods $A_{P+PD+CD}$, $A_{P+PD}$ and $A_{P+CD}$ exhibit increasing trends as the evolutionary process progresses. Notably, the method $A_{P}$, which does not incorporate any diversity metric (represented by the blue curve with $\blacklozenge$), exhibits a consistent decrease in HV values across generations. This observation suggests that incorporating at least one diversity metric contributes positively to overall performance in terms of $f_{P}$, $f_{PD}$, and $f_{CD}$. 

\begin{figure}[htbp]
    \centering
    \includegraphics[scale=0.39]{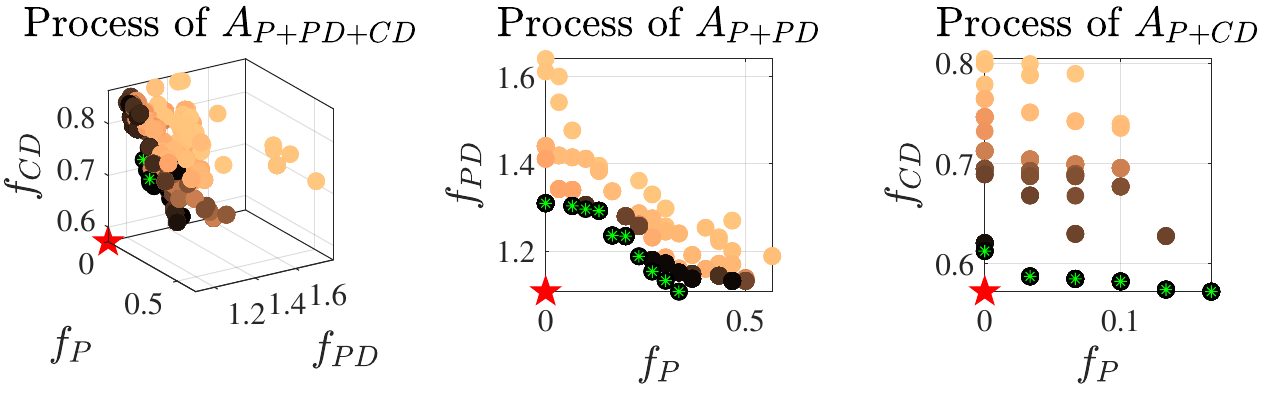}
    \caption{Non-dominated solutions of each generation in one arbitrary trial are plotted. Colours darken with the evolution progresses. Green stars highlight the non-dominated solutions in the final generation.}
    \label{fig:process}
\end{figure}

Fig. \ref{fig:process} illustrates the optimisation process of $A_{P+PD+CD}$, $A_{P+PD}$ and $A_{P+CD}$ by showcasing their respective optimisation objectives within arbitrarily selected trials. It's evident that the generators simultaneously approach the PFs (indicated by green stars) in terms of their optimised metrics, including playability and one or two diversity metrics. In other words, along the optimisation process, the solutions in the later stage are consistently better or at least no worse than those in the early stage in terms of all the optimised objectives. This visualisation indicates that our framework can simultaneously optimise multiple metrics, including various diversity metrics and playability, without sacrificing any single metric.

Moreover, the HV curve for $A_{P+PD+CD}$ (black curve with $\circ$) consistently ranks higher (better) than the others. This observation implies that $A_{P+PD+CD}$ achieves better tradeoffs among $f_{P}$, $f_{PD}$, and $f_{CD}$. Contrastingly, methods that solely consider a single diversity dimension, namely $A_{P+PD}$ (red curve with $\blacktriangle$) and $A_{P+CD}$ (green curve with $\times$), exhibit lower and nearly stagnant HV performance, suggesting limited effectiveness in terms of the three metrics. The HV values for the final generator set obtained by different methods across 30 trials are recorded in Table \ref{tab:HV_res}, providing a statistical analysis. The method $A_{P+PD+CD}$ exhibits superior performance with an average HV value of 0.5241, as determined by the Wilcoxon rank sum test at a 0.05 significance level. In contrast, both $A_{P+CD}$ and $A_{P+PD}$ exhibit lower HV values of 0.2770 and 0.3009, respectively. Notably, $A_{P}$ exhibits the lowest HV values at 0.0569, highlighting its focus solely on playability. These results statistically verify that our framework can effectively optimise multi-dimensional diversity in level generation.

\begin{table}[htbp]
  \centering
  \caption{HV and CPF of final generators averaged over 30 trials. ``+/$\approx$/-'' indicates that the averaged HV and CPF of the corresponding algorithm (specified by column header) is statistically better/similar/worse than the one of $A_{P+PD+CD}$ according to the Wilcoxon rank sum test with a 0.05 significance level. The standard deviations are in parentheses. The best averaged HV or CPF values are highlighted in bold.}
    \begin{tabular}{ccccc}
    \toprule
    & $A_{P+PD+CD}$ & $A_{P+CD}$ & $A_{P+PD}$&  $A_{P}$\\
    \midrule
     \thead{HV} & 
    \thead{\textbf{0.5241}\\\textbf{(1.86e-01)}} & \thead{0.2770\\(1.38e-01)-} & \thead{0.3009\\(6.80e-02)-} & \thead{0.0569\\(5.52e-02)-} \\
\midrule
     \thead{CPF} & 
    \thead{\textbf{0.0427}\\\textbf{(5.14e-02)}} & \thead{0.0196\\(3.62e-02)-} & \thead{0.0060\\(1.10e-02)-} & \thead{0.0022\\(8.31e-03)-} \\
    \bottomrule
    \end{tabular}%

  \label{tab:HV_res}%
\end{table}%

\paragraph{Distribution of tradeoffs}

\begin{figure*}[htbp]
    \centering
    \begin{subfloat}[\label{fig:case1_sub1}Overview of $f_{P}$-$f_{PD}$-$f_{CD}$.]{\includegraphics[width=.245\textwidth]{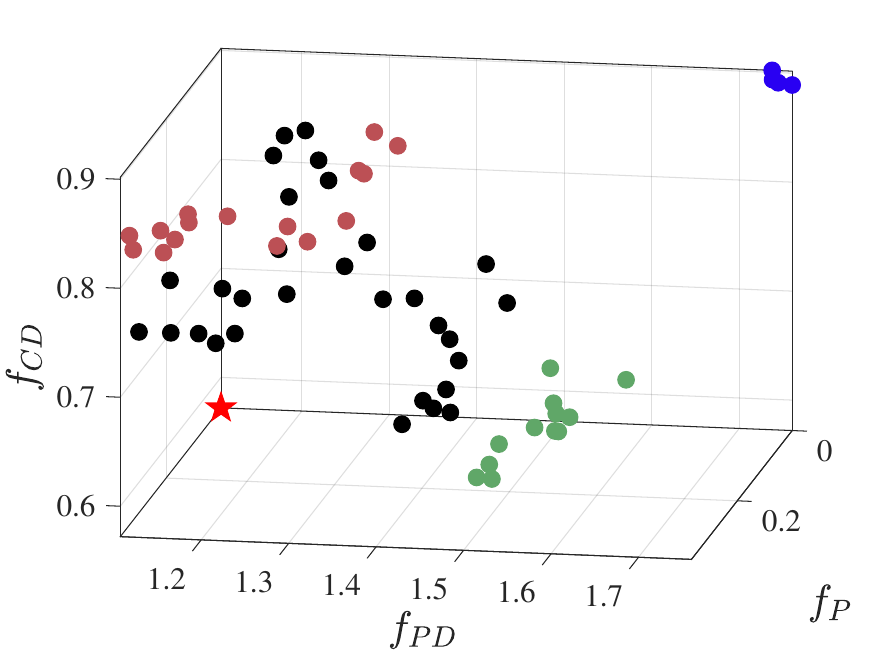}}
    \end{subfloat}
    \begin{subfloat}[\label{fig:case1_sub2}Perspective of $f_{P}$-$f_{PD}$.]{\includegraphics[width=.245\textwidth]{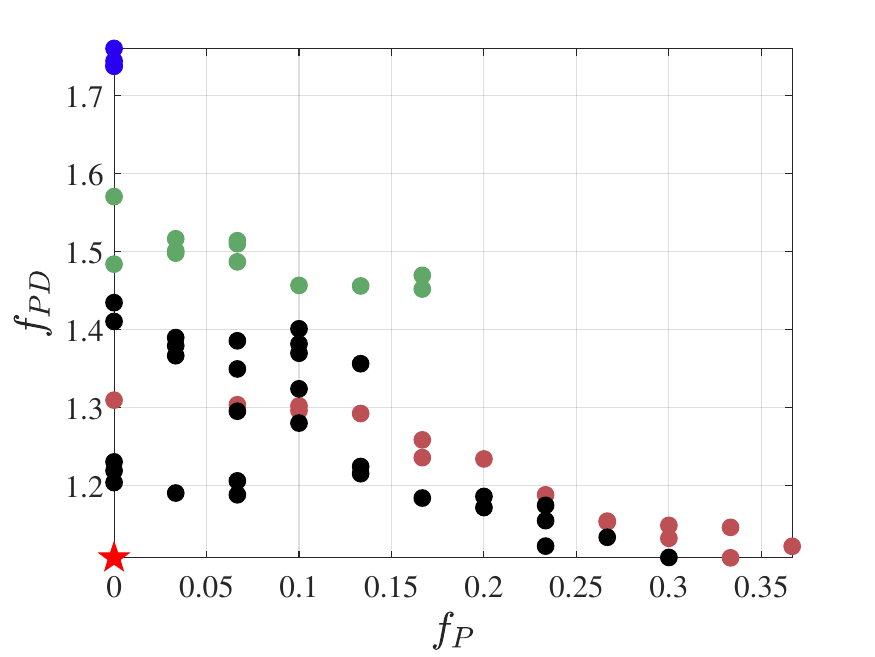}}
    \end{subfloat}
    \begin{subfloat}[\label{fig:case1_sub3}Perspective of $f_{P}$-$f_{CD}$.]{\includegraphics[width=.245\textwidth]{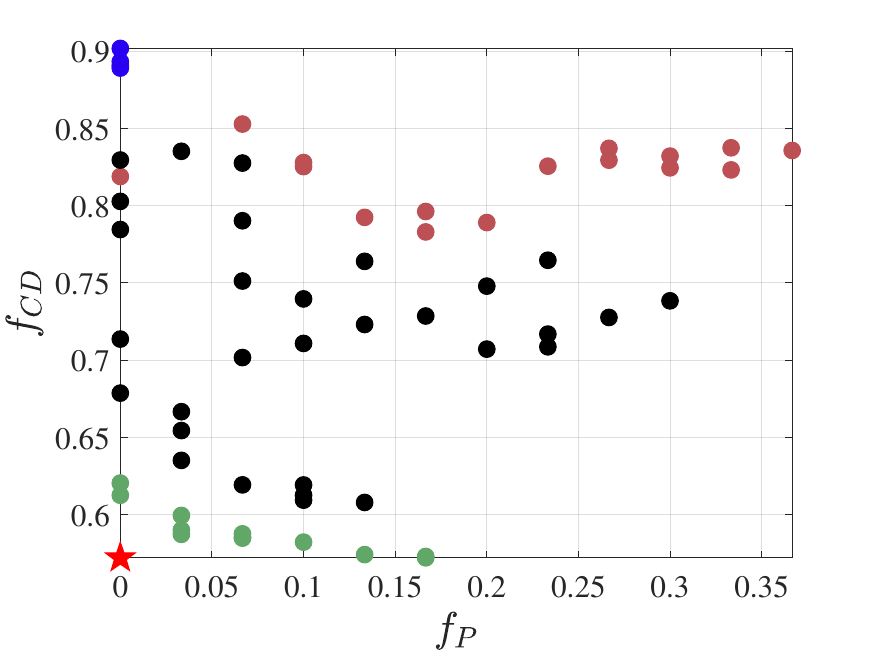}}
    \end{subfloat}
    \begin{subfloat}[\label{fig:case1_sub4}Perspective of $f_{PD}$-$f_{CD}$.]{\includegraphics[width=.245\textwidth]{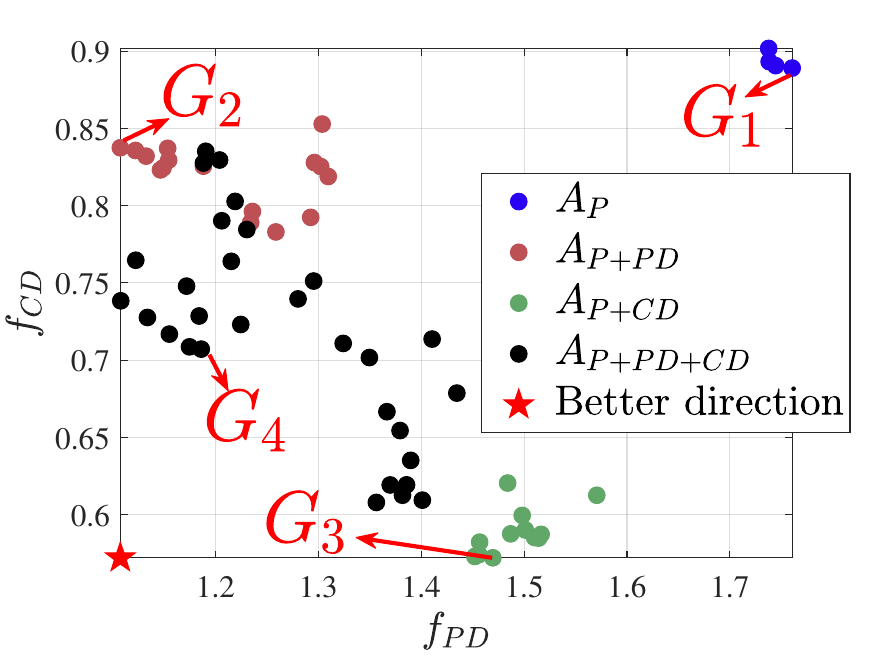}}
    \end{subfloat}
    \caption{Non-dominated solutions considering $f_{P}$, $f_{PD}$ and $f_{CD}$ obtained by $A_{P+PD+CD}$, $A_{P+PD}$, $A_{P+CD}$ and $A_{P}$, respectively, in a single run. Each plot offers a different coordinate perspective.}
    \label{fig:case1}
\end{figure*}  

\begin{figure*}[htbp]
    \centering
    \includegraphics[scale=0.575]{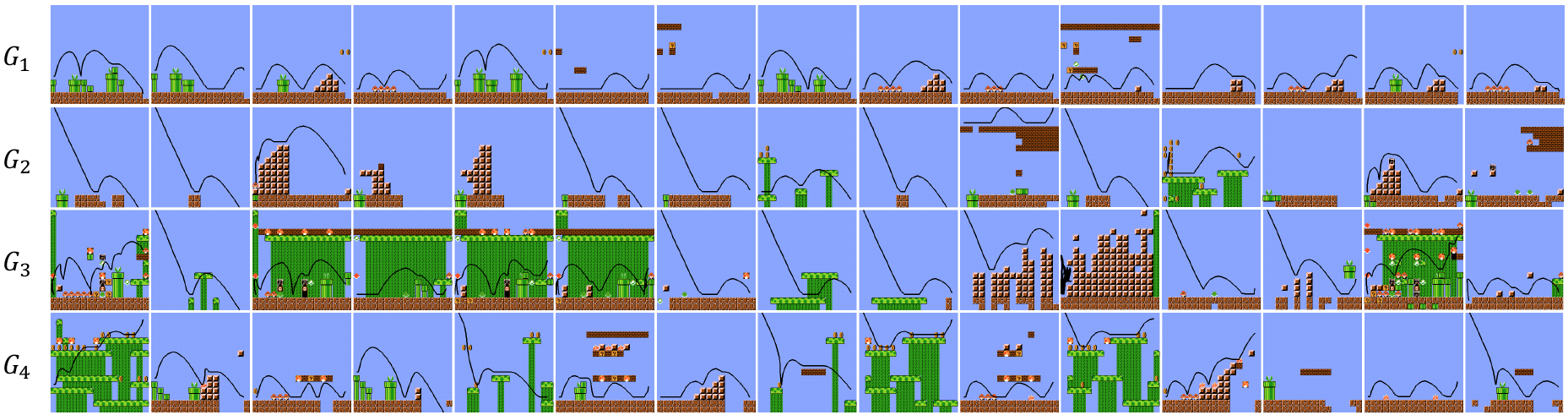}
    \caption{Level sampling examples of the generators $G_1$, $G_2$, $G_3$ and $G_4$ from Fig.~\ref{fig:case1_sub4}. The black trace indicates the agent's path.}
    \label{fig:case2}
\end{figure*}  

The generator sets, each representing a tradeoff, obtained by the four methods in the final generation of one randomly selected trial are shown in Fig. \ref{fig:case1}. Specifically, the non-dominated generators in relation to $f_{P}$, $f_{PD}$ and $f_{CD}$ across four different axes, including a 3D plot ($f_{P}$-$f_{PD}$-$f_{CD}$) and three 2D plots ($f_{P}$-$f_{PD}$, $f_{P}$-$f_{CD}$ and $f_{PD}$-$f_{CD}$), are presented. Furthermore, Fig. \ref{fig:case2} illustrates SMB game levels sampled from the four generators from Fig. \ref{fig:case1_sub4}, annotated by red arrows and texts in the $f_{PD}$-$f_{CD}$ coordinate plot since we more care about the performance of both diversity metrics. The black trace, indicating the playtrace, is simulated by the gameplaying agent of~\cite{zhang2022generating}.

Generators from $A_{P}$ (blue points) achieve the optimal performance for $f_{P}$ (0) but exhibit the worst performance for $f_{PD}$ (around 1.75) and $f_{CD}$ (around 0.9), as shown in the $f_{PD}$-$f_{CD}$ coordinate plot. For example, generator $G_1$ from $A_{P}$, also highlights that the levels sampled by $G_1$ are very similar in both $f_{CD}$ and $f_{PD}$. $A_{P+PD}$ (red points) performs less effectively than $A_{P+PD+CD}$ and $A_{P+CD}$ in $f_{CD}$, reaching only around 0.8. Generator $G_2$ from $A_{P+PD}$, indicates more diversity in agent traces but less diversity in level content, with some levels exhibiting similar patterns. Generators from $A_{P+CD}$ (green points) only achieve approximately 1.45 in $f_{PD}$. The levels generated by the $G_3$ from $A_{P+CD}$ contain more diverse elements, contributing to higher diversity in level $f_{CD}$ but the traces of $G_3$ demonstrate more similar patterns, resulting in a poorer $f_{PD}$. 

$A_{P+CD}$ (green points), which only optimises $f_{P}$ and $f_{CD}$, achieves better performance on $f_{CD}$ but worse performance on $f_{PD}$ compared to $A_{P+PD}$. However, $A_{P+PD}$ (red points) performs better on $f_{PD}$ and worse on $f_{CD}$. The observation suggests that the two diversity metrics, $f_{CD}$ and $f_{PD}$, conflict with each other. In other words, enhancing one diversity can negatively impact another. In contrast, $A_{P+PD+CD}$ can effectively optimise all three objectives. The generators from $A_{P+PD+CD}$ (black points) not only dominate those from $A_{P+PD}$ and $A_{P}$ but also show competitive performance with those from $A_{P+CD}$. The levels sampled by $G_4$ exhibit high diversity in terms of content and agent traces. 

Fig.~\ref{fig:case1} indicates that the solutions obtained by $A_{P+PD+CD}$ (black points) exhibit better diversity in terms of these objectives, demonstrating a balanced PF representation. Indeed, potential issues with an unbalanced representation of PF may be encountered when optimising other objectives utilising our framework. Even though our case study does not encounter this issue, existing methods, such as those in~\cite{10323115}, can be adapted to ensure a balanced PF representation.

\paragraph{Distribution of knee points}

\begin{figure}[htbp]
    \centering
    \includegraphics[scale=0.33]{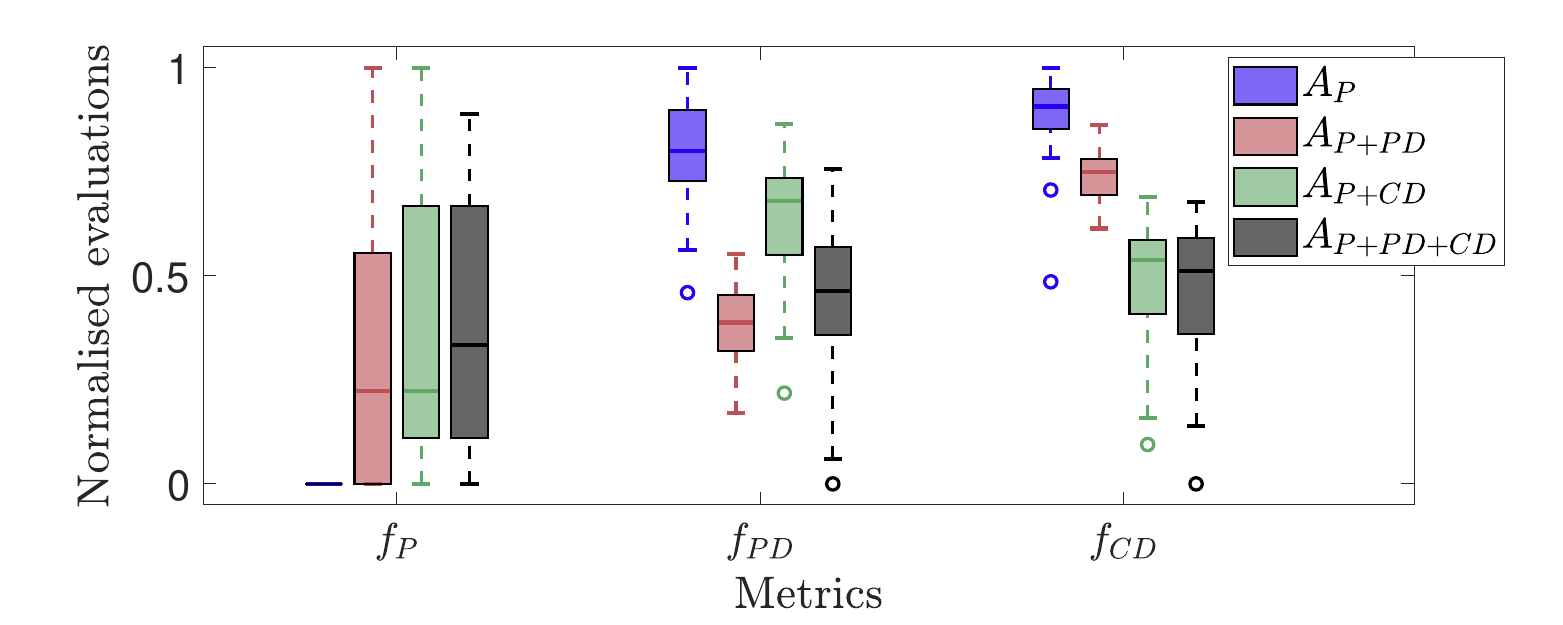}
    \caption{Boxplot showing the knee points obtained by $A_{P+PD+CD}$, $A_{P+PD}$, $A_{P+CD}$ and $A_{P}$, in terms of $f_{P}$, $f_{CD}$ and $f_{PD}$ over 30 trials. }
    \label{fig:knee_points}
\end{figure}

Fig. \ref{fig:knee_points} presents a boxplot of knee points obtained by the four comparison methods across 30 trials, considering $f_{P}$, $f_{CD}$, and $f_{PD}$. The widely used HV-based knee-point selection method~\cite{6975108} is used. In general, Fig. \ref{fig:knee_points} suggests that $A_{P+PD+CD}$ (black colour) exhibits a better balance among the three metrics.

$A_{P+PD+CD}$ ranks second in $f_{PD}$ and first in $f_{CD}$. Although $A_{P+PD+CD}$ ranks fourth in $f_{P}$, its performance remains comparable to that of other comparison methods. In contrast, $A_{P+PD}$ (red colour) and $A_{P+CD}$ (green colour) exhibit less effectively in $f_{CD}$ and $f_{PD}$, respectively, due to their failure to explicitly consider the respective objectives. Not surprisingly, $A_{P}$ (blue colour), which does not incorporate any diversity metrics, performs poorly in $f_{CD}$ and $f_{PD}$ but demonstrates better results in $f_{P}$.

In summary, our framework effectively demonstrates the capability to enhance multi-dimensional diversity metrics while maintaining good performance in playability.

\subsection{Effectiveness in Providing a Diverse Generator Set}

To assess the effectiveness of our framework in enhancing diversity within generator sets, we adopt two perspectives. First, we evaluate the diversity of the generator sets obtained by different methods utilising CPF. This analysis provides insights into the statistical extent of diversity achieved by each method. Second, we delve deeper into the implications of our framework by showcasing the distribution of diverse tradeoffs/generators within one randomly selected trail. Then, concrete examples of game levels are sampled and depicted by representative generators with varying tradeoffs, including those with the best and central rankings. By presenting these examples, we aim to provide a clearer understanding of the qualitative aspects of diversity enhancement.

\paragraph{CPF performance}
 
The CPF values for the final generator sets averaged across 30 trials from $A_{P+PD+CD}$, $A_{P+PD}$, $A_{P+CD}$ and $A_{P}$, are recorded in Table \ref{tab:HV_res}. $A_{P+PD+CD}$ exhibits the highest CPF value of 0.0427, outperforming other methods that consider fewer metrics. $A_{P+CD}$ and $A_{P+PD}$ rank second and third with values of 0.0196 and 0.0060, respectively. $A_{P}$, focusing solely on playability, performs the worst with a CPF value of 0.0022. These results suggest that incorporating all metrics $f_{P}$, $f_{PD}$, and $f_{CD}$ based on our framework results in better diversity across all three metrics.

\paragraph{Diverse tradeoffs}

\begin{figure*}[htbp]
    \centering
    \includegraphics[scale=0.275]{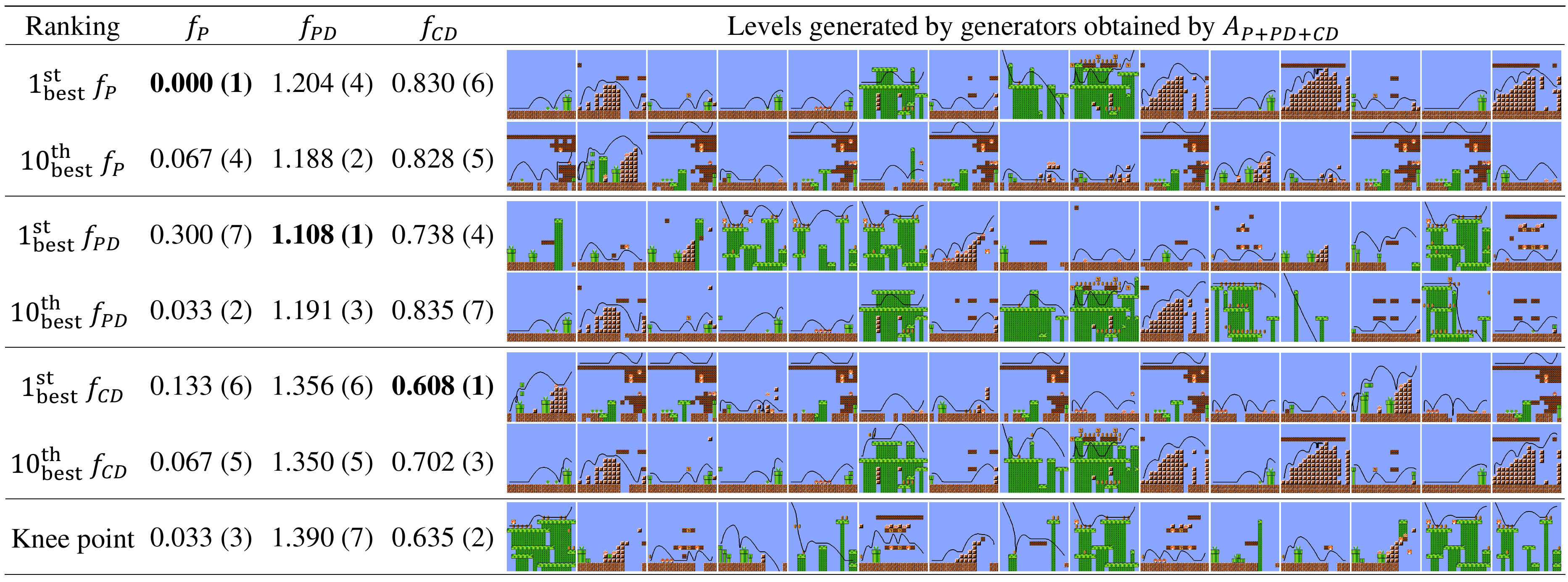}
    \caption{Metric values and 15 sampled levels for the best and 10th best generators for $f_{P}$, $f_{PD}$ and $f_{CD}$, respectively, in a single run of $A_{P+PD+CD}$, where the rankings among these generators are indicated in brackets.}
    \label{fig:tradeoffs}
\end{figure*} 

To demonstrate the diversity of tradeoffs among the three metrics ($f_{P}$, $f_{PD}$, and $f_{CD}$), we showcase the generator set obtained by $A_{P+PD+CD}$ in Fig. \ref{fig:case1}, as depicted in Fig. \ref{fig:tradeoffs}. Specifically, we plot the distribution of the non-dominated generators randomly chosen from a final generation in a single trial, as shown for $A_{P+PD+CD}$ in Fig. \ref{fig:case1}. Within these generators, seven generators are selected, including the knee point, as well as generators with the best and the 10th best performance according to each metric. Due to space limitations, only 15 out of the 30 levels used for evaluation are visualised from each selected generator, with the agent's playtrace indicated by the black curves. The evaluation values for each metric are recorded in the corresponding rows, with rankings among the seven generators indicated in parentheses. Additionally, we record the values of each metric in the corresponding rows, with rankings among the seven generators indicated in parentheses.

The generators from $A_{P+PD+CD}$ (black points) in Fig. \ref{fig:case1} exhibit significant diversity across $f_{P}$, $f_{PD}$, and $f_{CD}$. Fig. \ref{fig:case1} shows the extensive variability of these three metrics within $A_{P+PD+CD}$. In the $f_{PD}$-$f_{CD}$ plot, both $A_{P+PD}$ and $A_{P+CD}$ tend to converge to local regions, achieving around 0.8 in $f_{CD}$ or around 1.45 in $f_{PD}$, resulting in poor diversity. However, $A_{P+PD+CD}$ effectively bridges the gap between $A_{P+PD}$ and $A_{P+CD}$, offering a range of generators that span a broader spectrum of tradeoffs between $f_{PD}$ and $f_{CD}$.

Fig. \ref{fig:tradeoffs} illustrates the enhanced diversity of tradeoffs in terms of their metric values and sampled levels. The three extreme generators achieve significant performance in their respective metric, illustrating the inherent tradeoffs in optimising one metric over others. For instance, considering the comparison between the generator $1^{st}_{best}$ $f_{PD}$ (with $f_{PD}$ value of 1.108) and the generator $10^{th}_{best}$ $f_{PD}$ (with $f_{PD}$ value of 1.191), improving $f_{P}$ from 0.300 to 0.033 may lead to a slight compromise of $f_{PD}$ from 1.108 to 1.191. This capability enables decision-makers to select generators that best align with specific preferences and priorities, considering a range of scenarios and the diverse requirements of game players. Notably, comparing the generator $1^{st}_{best}$ $f_{P}$ obtained by $A_{P}$ in Fig.~\ref{fig:tradeoffs} with the generator $G_1$  obtained by $A_{P+PD+CD}$ in Fig.~\ref{fig:case2}, although both generators achieve the optimal playability, the generator $1^{st}_{best}$ $f_{P}$ can generate more diverse levels. This is because $A_{P+PD+CD}$ can consider and further enhance multi-dimensional diversity while optimising playability.

To summarise, our framework offers the capability of providing a diverse generator set, representing various tradeoffs among considered objectives in generating game levels. This diversity allows for the exploration of different combinations of objectives, enabling the creation of game levels that meet a wide range of metrics and preferences.

\section{Conclusion}\label{sec:conclusion}

This paper presents a framework to expand horizons of level diversity via MOEL. Our extensive experimental analysis validates the effectiveness of our framework in enhancing the multi-dimensional diversity of level generators and in providing a diverse set of generators. The models produced by our framework reflect varying tradeoffs among metrics, including playability and diversity. This assortment of generators enables decision-makers (e.g., designers) to select options that best align with specific preferences and priorities among these metrics~\cite{jaini2017trade}. Furthermore, our algorithm can provide a knee point generator, offering a ``central'' compromise among the considered metrics, even when decision-makers have no specific preferences~\cite{li2020does}. Additionally, we take the step to investigate the benefits of using MOEL to train generative models in PCG. The objectives in our work encompass playability and multi-dimensional diversity but are also applicable to other fields, such as affective modeling~\cite{10269152}. Moreover, while our study was conducted on a 2D platformer game, our principles and methods are applicable to both 2D and 3D game environments. This flexibility ensures that our contributions are relevant and valuable across various gaming genres and platforms.

Looking ahead, first, in addition to levels, other facets of content~\cite{Liapis2019Orchestrating}, such as music, narrative and social interaction, should be considered to foster deeper player engagement and enjoyment. Second, as the study~\cite{liy2024measuring} reviewed, there are many metrics for measuring the diversity of game scenarios. It would be beneficial to select a few key metrics~\cite{wang2016objective} that effectively capture all aspects of diversity. Third, we are also interested in optimising multiple diversity metrics using our framework to validate its scalability. Incorporating objective reduction methods~\cite{li2023offline} could be a good choice. Finally, the robustness and generalisation of our framework can be assessed by applying our approach to various game types~\cite{hu2024games} and other evaluation metrics~\cite{liy2024measuring}. Incorporating knowledge to adapt models trained on one game to generate content for new games is also worth exploration~\cite{Sarkar2024Procedural}, targeting to more general results rather than overfitting to a single scenario.

\section*{Acknowledgement}
Authors would like to thank all anonymous reviewers for their careful review and insightful comments.

\bibliographystyle{IEEEtran}
\bibliography{bibs/main_new}

\end{document}

%% file: tables/representation.tex
\begin{tabular}{cc|cc}
    \toprule
    Tile & Image & Tile & Image\\
   \midrule
    Solid & \includegraphics[width=4mm]{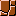} & Breakable & \includegraphics[width=4mm]{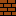} \\
    Empty &  & Jump-through platform & \includegraphics[width=4mm]{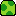} \\
    Pipe & \includegraphics[width=3.6mm]{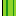}\includegraphics[width=3.6mm]{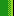}\includegraphics[width=3.6mm]{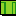}\includegraphics[width=3.6mm]{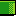} & Bullet Bill head & \includegraphics[width=4mm]{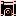} \\   
    Pipe with chomper & \includegraphics[width=3.6mm]{figures/tiles/a1.png}\includegraphics[width=3.6mm]{figures/tiles/a2.png}\includegraphics[width=3.6mm]{figures/tiles/TLP.png}\includegraphics[width=3.6mm]{figures/tiles/TRP.png} & Bullet Bill body & \includegraphics[width=4mm]{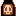} \\   
    Coin & \includegraphics[width=4mm]{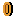} & Bullet Bill Coin box & \includegraphics[width=4mm]{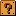} \\           Mushroom box & \includegraphics[width=4mm]{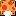} & Mushroom block & \includegraphics[width=4mm]{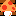} \\      
    Bonus life box & \includegraphics[width=4mm]{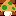} & Invisible bonus life box & \includegraphics[width=4mm]{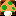} \\    
    Invisible coin box & \includegraphics[width=4mm]{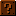} & Goomba & \includegraphics[width=4mm]{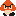} \\    
    Green Koopa & \includegraphics[width=4mm]{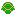} & Red Koopa & \includegraphics[width=4mm]{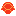} \\    
    Flying green Koopa & \includegraphics[width=4mm]{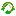} & Flying Red Koopa & \includegraphics[width=4mm]{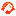} \\    
    Spiky & \includegraphics[width=4mm]{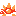} & &  \\    
    \bottomrule
\end{tabular}

%% file: main-CameraReady.bbl
\begin{thebibliography}{10}
\providecommand{\url}[1]{#1}
\csname url@samestyle\endcsname
\providecommand{\newblock}{\relax}
\providecommand{\bibinfo}[2]{#2}
\providecommand{\BIBentrySTDinterwordspacing}{\spaceskip=0pt\relax}
\providecommand{\BIBentryALTinterwordstretchfactor}{4}
\providecommand{\BIBentryALTinterwordspacing}{\spaceskip=\fontdimen2\font plus
\BIBentryALTinterwordstretchfactor\fontdimen3\font minus \fontdimen4\font\relax}
\providecommand{\BIBforeignlanguage}[2]{{%
\expandafter\ifx\csname l@#1\endcsname\relax
\typeout{** WARNING: IEEEtran.bst: No hyphenation pattern has been}%
\typeout{** loaded for the language `#1'. Using the pattern for}%
\typeout{** the default language instead.}%
\else
\language=\csname l@#1\endcsname
\fi
#2}}
\providecommand{\BIBdecl}{\relax}
\BIBdecl

\bibitem{shaker2016procedural}
N.~Shaker, J.~Togelius, and M.~J. Nelson, \emph{Procedural content generation in games}.\hskip 1em plus 0.5em minus 0.4em\relax Springer, 2016.

\bibitem{liu2021deep}
J.~Liu, S.~Snodgrass, A.~Khalifa, S.~Risi, G.~N. Yannakakis, and J.~Togelius, ``Deep learning for procedural content generation,'' \emph{Neural Computing and Applications}, vol.~33, no.~1, pp. 19--37, 2021.

\bibitem{guzdial2022procedural}
M.~Guzdial, S.~Snodgrass, and A.~J. Summerville, \emph{Procedural Content Generation Via Machine Learning: An Overview}.\hskip 1em plus 0.5em minus 0.4em\relax Springer, 2022.

\bibitem{10.1145/3422622}
I.~Goodfellow, J.~Pouget-Abadie, M.~Mirza, B.~Xu, D.~Warde-Farley, S.~Ozair, A.~Courville, and Y.~Bengio, ``Generative adversarial networks,'' \emph{Communications of the ACM}, vol.~63, no.~11, p. 139–144, 2020.

\bibitem{DBLP:journals/corr/KingmaW13}
D.~P. Kingma and M.~Welling, ``Auto-encoding variational bayes,'' in \emph{2nd International Conference on Learning Representations, {ICLR}}, 2014.

\bibitem{volz2018evolving}
V.~Volz, J.~Schrum, J.~Liu, S.~M. Lucas, A.~Smith, and S.~Risi, ``Evolving {Mario} levels in the latent space of a deep convolutional generative adversarial network,'' in \emph{Proceedings of the Genetic and Evolutionary Computation Conference}.\hskip 1em plus 0.5em minus 0.4em\relax ACM, 2018, pp. 221--228.

\bibitem{snodgrass2020multi}
S.~Snodgrass and A.~Sarkar, ``Multi-domain level generation and blending with sketches via example-driven {BSP} and variational autoencoders,'' in \emph{Proceedings of the 15th International Conference on the Foundations of Digital Games}.\hskip 1em plus 0.5em minus 0.4em\relax ACM, 2020, pp. 1--11.

\bibitem{yannakakis2018artificial}
G.~N. Yannakakis and J.~Togelius, \emph{Artificial Intelligence and Games}.\hskip 1em plus 0.5em minus 0.4em\relax Springer, 2018.

\bibitem{Liapis2019Orchestrating}
A.~Liapis, G.~N. Yannakakis, M.~J. Nelson, M.~Preuss, and R.~Bidarra, ``Orchestrating {Game} {Generation},'' \emph{IEEE Transactions on Games}, vol.~11, no.~1, pp. 48--68, 2019.

\bibitem{gravina2019procedural}
D.~Gravina, A.~Khalifa, A.~Liapis, J.~Togelius, and G.~N. Yannakakis, ``Procedural content generation through quality diversity,'' in \emph{2019 IEEE Conference on Games}.\hskip 1em plus 0.5em minus 0.4em\relax IEEE, 2019, pp. 1--8.

\bibitem{Jordanous2011Evaluating}
A.~Jordanous, ``Evaluating evaluation: Assessing progress in computational creativity research,'' in \emph{Proceedings of the 2nd International Conference on Computational Creativity}, 2011, pp. 102--107.

\bibitem{liy2024measuring}
Y.~Li, Z.~Wang, Q.~Zhang, and J.~Liu, ``Measuring diversity of game scenarios,'' \emph{arXiv preprint arXiv:2404.15192}, 2024.

\bibitem{shu2021experience}
T.~Shu, J.~Liu, and G.~N. Yannakakis, ``Experience-driven {PCG} via reinforcement learning: A {Super Mario Bros} study,'' in \emph{2021 IEEE Conference on Games}.\hskip 1em plus 0.5em minus 0.4em\relax IEEE, 2021, pp. 1--9.

\bibitem{wang2022fun}
Z.~Wang, J.~Liu, and G.~N. Yannakakis, ``The fun facets of {Mario}: Multifaceted experience-driven {PCG} via reinforcement learning,'' in \emph{International Conference on the Foundations of Digital Games}, 2022, pp. 1--8.

\bibitem{Liapis2015Constrained}
A.~Liapis, G.~N. Yannakakis, and J.~Togelius, ``Constrained novelty search: A study on game content generation,'' \emph{Evolutionary Computation}, vol.~23, no.~1, pp. 101--129, 2015.

\bibitem{fontaine2021illuminating}
M.~C. Fontaine, R.~Liu, A.~Khalifa, J.~Modi, J.~Togelius, A.~K. Hoover, and S.~Nikolaidis, ``Illuminating {M}ario scenes in the latent space of a generative adversarial network,'' in \emph{Proceedings of the AAAI Conference on Artificial Intelligence}, vol.~35, no.~7, 2021, pp. 5922--5930.

\bibitem{bontrager2021learning}
P.~Bontrager and J.~Togelius, ``Learning to generate levels from nothing,'' in \emph{2021 IEEE Conference on Games}, 2021, pp. 1--8.

\bibitem{preuss2014searching}
M.~Preuss, A.~Liapis, and J.~Togelius, ``Searching for good and diverse game levels,'' in \emph{2014 IEEE Conference on Computational Intelligence and Games}, 2014, pp. 1--8.

\bibitem{nam2019generation}
S.~Nam and K.~Ikeda, ``Generation of diverse stages in turn-based role-playing game using reinforcement learning,'' in \emph{2019 IEEE Conference on Games}.\hskip 1em plus 0.5em minus 0.4em\relax IEEE, 2019, pp. 1--8.

\bibitem{beukman2022procedural}
M.~Beukman, C.~W. Cleghorn, and S.~James, ``Procedural content generation using neuroevolution and novelty search for diverse video game levels,'' in \emph{Genetic and Evolutionary Computation Conference}.\hskip 1em plus 0.5em minus 0.4em\relax ACM, 2022, pp. 1028--1037.

\bibitem{lehman2011abandoning}
J.~Lehman and K.~O. Stanley, ``Abandoning objectives: Evolution through the search for novelty alone,'' \emph{Evolutionary Computation}, vol.~19, no.~2, pp. 189--223, 2011.

\bibitem{sudhakaran2024mariogpt}
S.~Sudhakaran, M.~Gonz{\'a}lez-Duque, M.~Freiberger, C.~Glanois, E.~Najarro, and S.~Risi, ``{MarioGPT}: Open-ended text2level generation through large language models,'' \emph{Advances in Neural Information Processing Systems}, vol.~36, pp. 1--13, 2024.

\bibitem{Sarkar2024Procedural}
A.~Sarkar, M.~Guzdial, S.~Snodgrass, A.~Summerville, T.~Machado, and G.~Smith, ``Procedural content generation via knowledge transformation ({PCG-KT}),'' \emph{IEEE Transactions on Games}, vol.~16, no.~1, pp. 36--50, 2024.

\bibitem{AwiSch2023worldgan}
M.~Awiszus, F.~Schubert, and B.~Rosenhahn, ``Wor(l)d-{GAN}: Toward natural-language-based {PCG} in {Minecraft},'' \emph{IEEE Transactions on Games}, vol.~15, no.~2, pp. 182--192, 2023.

\bibitem{torrado2020bootstrapping}
R.~Rodriguez~Torrado, A.~Khalifa, M.~Cerny~Green, N.~Justesen, S.~Risi, and J.~Togelius, ``Bootstrapping conditional {GAN}s for video game level generation,'' in \emph{2020 IEEE Conference on Games}.\hskip 1em plus 0.5em minus 0.4em\relax IEEE, 2020, pp. 41--48.

\bibitem{li2019quality}
M.~Li and X.~Yao, ``Quality evaluation of solution sets in multiobjective optimisation: A survey,'' \emph{ACM Computing Surveys (CSUR)}, vol.~52, no.~2, pp. 1--38, 2019.

\bibitem{ZHOU201132}
A.~Zhou, B.-Y. Qu, H.~Li, S.-Z. Zhao, P.~N. Suganthan, and Q.~Zhang, ``Multiobjective evolutionary algorithms: A survey of the state of the art,'' \emph{Swarm and Evolutionary Computation}, vol.~1, no.~1, pp. 32--49, 2011.

\bibitem{10.1145/2792984}
B.~Li, J.~Li, K.~Tang, and X.~Yao, ``Many-objective evolutionary algorithms: A survey,'' \emph{ACM Computing Surveys (CSUR)}, vol.~48, no.~1, pp. 1--35, 2015.

\bibitem{lara2014balance}
R.~Lara-Cabrera, C.~Cotta, and A.~J. Fern{\'a}ndez-Leiva, ``On balance and dynamism in procedural content generation with self-adaptive evolutionary algorithms,'' \emph{Natural Computing}, vol.~13, pp. 157--168, 2014.

\bibitem{ruela2020multi}
A.~S. Ruela, K.~V. Delgado, and J.~Bernardes, ``A multi-objective evolutionary approach for the nonlinear scale-free level problem,'' \emph{Applied Intelligence}, vol.~50, pp. 4223--4240, 2020.

\bibitem{zhang2024interpreting}
Q.~Zhang, Y.~Li, Y.~Lin, H.~Wang, and J.~Liu, ``Interpreting multi-objective evolutionary algorithms via {Sokoban} level generation,'' in \emph{2024 IEEE Conference on Games}, 2024, pp. 1--2.

\bibitem{wang2014two_arch2}
H.~Wang, L.~Jiao, and X.~Yao, ``Two\_arch2: An improved two-archive algorithm for many-objective optimization,'' \emph{IEEE Transactions on Evolutionary Computation}, vol.~19, no.~4, pp. 524--541, 2014.

\bibitem{deb2002fast}
K.~Deb, A.~Pratap, S.~Agarwal, and T.~Meyarivan, ``A fast and elitist multiobjective genetic algorithm: {NSGA-II},'' \emph{IEEE Transactions on Evolutionary Computation}, vol.~6, no.~2, pp. 182--197, 2002.

\bibitem{khalifa2020multi}
A.~Khalifa and J.~Togelius, ``Multi-objective level generator generation with {Marahel},'' in \emph{International Conference on the Foundations of Digital Games}, 2020, pp. 1--8.

\bibitem{kutzias2023recent}
D.~Kutzias and S.~von Mammen, ``Recent advances in procedural generation of buildings: From diversity to integration,'' \emph{IEEE Transactions on Games}, pp. 16--35, 2023.

\bibitem{withington2023right}
O.~Withington and L.~Tokarchuk, ``The right variety: Improving expressive range analysis with metric selection methods,'' in \emph{International Conference on the Foundations of Digital Games}, 2023, pp. 1--11.

\bibitem{Shaker2016Evaluating}
N.~Shaker, G.~Smith, and G.~N. Yannakakis, \emph{Evaluating content generators}.\hskip 1em plus 0.5em minus 0.4em\relax Springer, 2016, pp. 215--224.

\bibitem{zakaria2023procedural}
Y.~Zakaria, M.~Fayek, and M.~Hadhoud, ``Procedural level generation for {Sokoban} via deep learning: An experimental study,'' \emph{IEEE Transactions on Games}, vol.~15, no.~1, pp. 108--120, 2023.

\bibitem{earle2021learning}
S.~Earle, M.~Edwards, A.~Khalifa, P.~Bontrager, and J.~Togelius, ``Learning controllable content generators,'' in \emph{2021 IEEE Conference on Games}.\hskip 1em plus 0.5em minus 0.4em\relax IEEE, 2021, pp. 1--9.

\bibitem{marino2015empirical}
J.~Mari{\~n}o, W.~Reis, and L.~Lelis, ``An empirical evaluation of evaluation metrics of procedurally generated {Mario} levels,'' in \emph{Conference on Artificial Intelligence and Interactive Digital Entertainment}, vol.~11, no.~1.\hskip 1em plus 0.5em minus 0.4em\relax AAAI, 2015, pp. 44--50.

\bibitem{lucas2019tile}
S.~M. Lucas and V.~Volz, ``Tile pattern {KL}-divergence for analysing and evolving game levels,'' in \emph{Proceedings of the Genetic and Evolutionary Computation Conference}.\hskip 1em plus 0.5em minus 0.4em\relax ACM, 2019, pp. 170--178.

\bibitem{berndt1994using}
D.~J. Berndt and J.~Clifford, ``Using dynamic time warping to find patterns in time series.'' in \emph{Proceedings of the 3rd International Conference on Knowledge Discovery and Data Mining}, vol.~10, no.~16, 1994, pp. 359--370.

\bibitem{liapis2013enhancements}
A.~Liapis, G.~N. Yannakakis, and J.~Togelius, ``Enhancements to constrained novelty search: Two-population novelty search for generating game content,'' in \emph{Proceedings of the 15th Annual Conference on Genetic and Evolutionary Computation}, 2013, pp. 343--350.

\bibitem{medina2023evolving}
A.~Medina, M.~Richey, M.~Mueller, and J.~Schrum, ``Evolving flying machines in {Minecraft} using quality diversity,'' in \emph{Proceedings of the Genetic and Evolutionary Computation Conference}.\hskip 1em plus 0.5em minus 0.4em\relax ACM, 2023, pp. 1418--1426.

\bibitem{lai2022mixed}
G.~Lai, F.~F. Leymarie, and W.~Latham, ``On mixed-initiative content creation for video games,'' \emph{IEEE Transactions on Games}, vol.~14, no.~4, pp. 543--557, 2022.

\bibitem{Mitigating_unfairness_2023}
Q.~Zhang, J.~Liu, Z.~Zhang, J.~Wen, B.~Mao, and X.~Yao, ``Mitigating unfairness via evolutionary multiobjective ensemble learning,'' \emph{IEEE Transactions on Evolutionary Computation}, vol.~27, no.~4, pp. 848--862, 2023.

\bibitem{fairerML_2021}
------, ``Fairer machine learning through multi-objective evolutionary learning,'' in \emph{International Conference on Artificial Neural Networks}.\hskip 1em plus 0.5em minus 0.4em\relax Springer, 2021, pp. 111--123.

\bibitem{togelius2011search}
J.~Togelius, G.~N. Yannakakis, K.~O. Stanley, and C.~Browne, ``Search-based procedural content generation: A taxonomy and survey,'' \emph{IEEE Transactions on Computational Intelligence and AI in Games}, vol.~3, no.~3, pp. 172--186, 2011.

\bibitem{summerville2018procedural}
A.~Summerville, S.~Snodgrass, M.~Guzdial, C.~Holmg{\aa}rd, A.~K. Hoover, A.~Isaksen, A.~Nealen, and J.~Togelius, ``Procedural content generation via machine learning ({PCGML}),'' \emph{IEEE Transactions on Games}, vol.~10, no.~3, pp. 257--270, 2018.

\bibitem{hu2024games}
C.~Hu, Y.~Zhao, Z.~Wang, H.~Du, and J.~Liu, ``Games for artificial intelligence research: A review and perspectives,'' \emph{IEEE Transactions on Artificial Intelligence}, pp. 1--20, 2024.

\bibitem{zhang2019self}
H.~Zhang, I.~Goodfellow, D.~Metaxas, and A.~Odena, ``Self-attention generative adversarial networks,'' in \emph{International Conference on Machine Learning}.\hskip 1em plus 0.5em minus 0.4em\relax PMLR, 2019, pp. 7354--7363.

\bibitem{wang2024negatively}
\BIBentryALTinterwordspacing
Z.~Wang, C.~Hu, J.~Liu, and X.~Yao, ``Negatively correlated ensemble reinforcement learning for online diverse game level generation,'' in \emph{International Conference on Learning Representations}, 2024. [Online]. Available: \url{https://openreview.net/forum?id=iAW2EQXfwb}
\BIBentrySTDinterwordspacing

\bibitem{wang2019evolutionary}
C.~Wang, C.~Xu, X.~Yao, and D.~Tao, ``Evolutionary generative adversarial networks,'' \emph{IEEE Transactions on Evolutionary Computation}, vol.~23, no.~6, pp. 921--934, 2019.

\bibitem{MOGAN2020}
M.~Baioletti, C.~A.~C. Coello, G.~Di~Bari, and V.~Poggioni, ``Multi-objective evolutionary {GAN},'' in \emph{Proceedings of the 2020 Genetic and Evolutionary Computation Conference Companion}.\hskip 1em plus 0.5em minus 0.4em\relax ACM, 2020, pp. 1824--1831.

\bibitem{khalifa2020pcgrl}
A.~Khalifa, P.~Bontrager, S.~Earle, and J.~Togelius, ``{PCGRL}: Procedural content generation via reinforcement learning,'' \emph{Proceedings of the Sixteenth AAAI Conference on Artificial Intelligence and Interactive Digital Entertainment}, vol.~16, no.~1, pp. 95--101, 2020.

\bibitem{8456559}
Y.~Jin, H.~Wang, T.~Chugh, D.~Guo, and K.~Miettinen, ``Data-driven evolutionary optimization: An overview and case studies,'' \emph{IEEE Transactions on Evolutionary Computation}, vol.~23, no.~3, pp. 442--458, 2019.

\bibitem{SDEPMOEA_2023}
Q.~Zhang, J.~Liu, and X.~Yao, ``An efficient many objective optimization algorithm with few parameters,'' \emph{Swarm and Evolutionary Computation}, vol.~83, p. 101405, 2023.

\bibitem{karakovskiy2012mario}
S.~Karakovskiy and J.~Togelius, ``The {Mario AI} benchmark and competitions,'' \emph{IEEE Transactions on Computational Intelligence and AI in Games}, vol.~4, no.~1, pp. 55--67, 2012.

\bibitem{togelius2013procedural}
J.~Togelius, A.~J. Champandard, P.~L. Lanzi, M.~Mateas, A.~Paiva, M.~Preuss, and K.~O. Stanley, ``Procedural content generation: Goals, challenges and actionable steps,'' in \emph{Artificial and Computational Intelligence in Games}.\hskip 1em plus 0.5em minus 0.4em\relax Schloss Dagstuhl--Leibniz-Zentrum fuer Informatik, 2013, vol.~6, pp. 61--75.

\bibitem{awiszus2020toad}
M.~Awiszus, F.~Schubert, and B.~Rosenhahn, ``{TOAD-GAN}: coherent style level generation from a single example,'' in \emph{Proceedings of the AAAI Conference on Artificial Intelligence and Interactive Digital Entertainment}, vol.~16, no.~1, 2020, pp. 10--16.

\bibitem{zhang2022generating}
K.~Zhang, J.~Bai, and J.~Liu, ``Generating game levels of diverse behaviour engagement,'' in \emph{2022 IEEE Conference on Games}.\hskip 1em plus 0.5em minus 0.4em\relax IEEE, 2022, pp. 167--174.

\bibitem{Shang2021Survey}
K.~Shang, H.~Ishibuchi, L.~He, and L.~M. Pang, ``A survey on the hypervolume indicator in evolutionary multiobjective optimization,'' \emph{IEEE Transactions on Evolutionary Computation}, vol.~25, no.~1, pp. 1--20, 2021.

\bibitem{Tian2019Diversity}
Y.~Tian, R.~Cheng, X.~Zhang, M.~Li, and Y.~Jin, ``Diversity assessment of multi-objective evolutionary algorithms: Performance metric and benchmark problems,'' \emph{IEEE Computational Intelligence Magazine}, vol.~14, no.~3, pp. 61--74, 2019.

\bibitem{6975108}
X.~Zhang, Y.~Tian, and Y.~Jin, ``A knee point-driven evolutionary algorithm for many-objective optimization,'' \emph{IEEE Transactions on Evolutionary Computation}, vol.~19, no.~6, pp. 761--776, 2015.

\bibitem{das1999characterizing}
I.~Das, ``On characterizing the “knee” of the {Pareto} curve based on normal-boundary intersection,'' \emph{Structural Optimization}, vol.~18, pp. 107--115, 1999.

\bibitem{10323115}
Z.~Song, H.~Wang, B.~Xue, and M.~Zhang, ``Balancing different optimization difficulty between objectives in multi-objective feature selection,'' \emph{IEEE Transactions on Evolutionary Computation}, pp. 1--14, 2023.

\bibitem{jaini2017trade}
N.~Jaini and S.~Utyuzhnikov, ``Trade-off ranking method for multi-criteria decision analysis,'' \emph{Journal of Multi-Criteria Decision Analysis}, vol.~24, no. 3-4, pp. 121--132, 2017.

\bibitem{li2020does}
K.~Li, M.~Liao, K.~Deb, G.~Min, and X.~Yao, ``Does preference always help? a holistic study on preference-based evolutionary multiobjective optimization using reference points,'' \emph{IEEE Transactions on Evolutionary Computation}, vol.~24, no.~6, pp. 1078--1096, 2020.

\bibitem{10269152}
G.~N. Yannakakis and D.~Melhart, ``Affective game computing: A survey,'' \emph{Proceedings of the IEEE}, vol. 111, no.~10, pp. 1423--1444, 2023.

\bibitem{wang2016objective}
H.~Wang and X.~Yao, ``Objective reduction based on nonlinear correlation information entropy,'' \emph{Soft Computing}, vol.~20, pp. 2393--2407, 2016.

\bibitem{li2023offline}
G.~Li, Z.~Wang, Q.~Zhang, and J.~Sun, ``Offline and online objective reduction via {Gaussian} mixture model clustering,'' \emph{IEEE Transactions on Evolutionary Computation}, vol.~27, no.~2, pp. 341--354, 2023.

\end{thebibliography}
